\def\year{2024}\relax
\documentclass[letterpaper]{article} 
\usepackage{aaai24}  
\usepackage{times}  
\usepackage{helvet}  
\usepackage{courier}  
\usepackage[hyphens]{url}  
\usepackage{graphicx} 
\usepackage{makecell}
\usepackage{rotating}
\usepackage{multirow}
\usepackage{booktabs}
\usepackage{amsmath}

\usepackage{natbib}  
\usepackage{caption} 
\DeclareCaptionStyle{ruled}{labelfont=normalfont,labelsep=colon,strut=off} 
\usepackage{algorithm}
\usepackage{algorithmic}
\usepackage{multirow}
\usepackage{subcaption}
\usepackage{tikz}
\usepackage{lipsum}
\usepackage{paralist}
\usepackage{color}
\newcommand*\circled[1]{\tikz[baseline=(char.base)]{
\node[shape=circle,draw=black,fill=white,inner sep=0.6pt] (char) {\textcolor{black}{\footnotesize \textbf{#1}}};}}
\DeclareRobustCommand\onedot{\futurelet\.}
\def\@onedot{\ifx\.\else.\null\fi\xspace}

\def\eg{\emph{e.g.}\onedot} 
\def\ie{\emph{i.e.}\onedot} 
 
 \def\vs{\emph{vs}\onedot}
 
\def\etal{\emph{et al}\onedot}
\usepackage{subcaption}

\makeatother
\usepackage{newfloat}
\usepackage{listings}
\floatstyle{ruled}
\newfloat{listing}{tb}{lst}{}
\floatname{listing}{Listing}

\usepackage{xcolor}

\definecolor{generate-color}{RGB}{80,0,80}

\usepackage{caption}
\usepackage{subcaption}
\usepackage{amsfonts}
\usepackage{color, colortbl}

\definecolor{LightCyan}{rgb}{0.88,1,1}
\title{Navigating Open Set Scenarios for Skeleton-based Action Recognition}
\author{
Kunyu Peng\textsuperscript{\rm 1},
Cheng Yin\textsuperscript{\rm 1},
Junwei Zheng\textsuperscript{\rm 1},
Ruiping Liu\textsuperscript{\rm 1},
David Schneider\textsuperscript{\rm 1},
Jiaming Zhang\textsuperscript{\rm 1},\\Kailun Yang\textsuperscript{\rm 2,}\thanks{Corresponding author.},
M. Saquib Sarfraz\textsuperscript{\rm 1,3},
Rainer Stiefelhagen\textsuperscript{\rm 1},
Alina Roitberg\textsuperscript{\rm 4}
}
\affiliations{
    \textsuperscript{\rm 1}Insitute for Anthropomatics and Robotics, Karlsruhe Institute of Technology, Germany\\
    \textsuperscript{\rm 2}School of Robotics, Hunan University, China\\
    \textsuperscript{\rm 3}Mercedes-Benz Tech Innovation, Germany\\
    \textsuperscript{\rm 4}Institute for Artificial Intelligence, University of Stuttgart, Germany\\
    firstname.lastname@kit.edu,~kailun.yang@hnu.edu.cn,~alina.roitberg@ki.uni-stuttgart.de
}

\usepackage{bibentry}

\begin{document}

\maketitle

\begin{abstract}
In real-world scenarios, human actions often fall outside the distribution of training data, making it crucial for models to recognize known actions and reject unknown ones. However, using pure skeleton data in such open-set conditions poses challenges due to the lack of visual background cues and the distinct sparse structure of body pose sequences. In this paper, we tackle the unexplored \textbf{O}pen-\textbf{S}et \textbf{S}keleton-based \textbf{A}ction \textbf{R}ecognition (OS-SAR) task and formalize the benchmark on three skeleton-based datasets. We assess the performance of seven established open-set approaches on our task and identify their limits and critical generalization issues when dealing with skeleton information. To address these challenges, we propose a distance-based cross-modality ensemble method that leverages the cross-modal alignment of skeleton joints, bones, and velocities to achieve superior open-set recognition performance. We refer to the key idea as CrossMax - an approach that utilizes a novel cross-modality mean max discrepancy suppression mechanism to align latent spaces during training and a cross-modality distance-based logits refinement method during testing. CrossMax outperforms existing approaches and consistently yields state-of-the-art results across all datasets and backbones. The benchmark, code, and models will be released at https://github.com/KPeng9510/OS-SAR.\footnote{This work was supported in part by the SmartAge project sponsored by the Carl Zeiss Stiftung (P2019-01-003; 2021-2026), the University of Excellence through the ``KIT Future Fields'' project, in part by the Helmholtz Association Initiative and Networking Fund on the HoreKA@KIT partition and the state of Baden-Württemberg through bwHPC and the German Research Foundation (DFG) through grant INST 35/1597-1 FUGG. A. Roitberg was supported by the Deutsche Forschungsgemeinschaft (DFG) under Germany’s Excellence Strategy - EXC 2075.}
\end{abstract}

\section{Introduction}

\begin{figure}[t]
    \centering
    \begin{minipage}[t]{0.45\columnwidth}
        \includegraphics[width=0.99\linewidth]{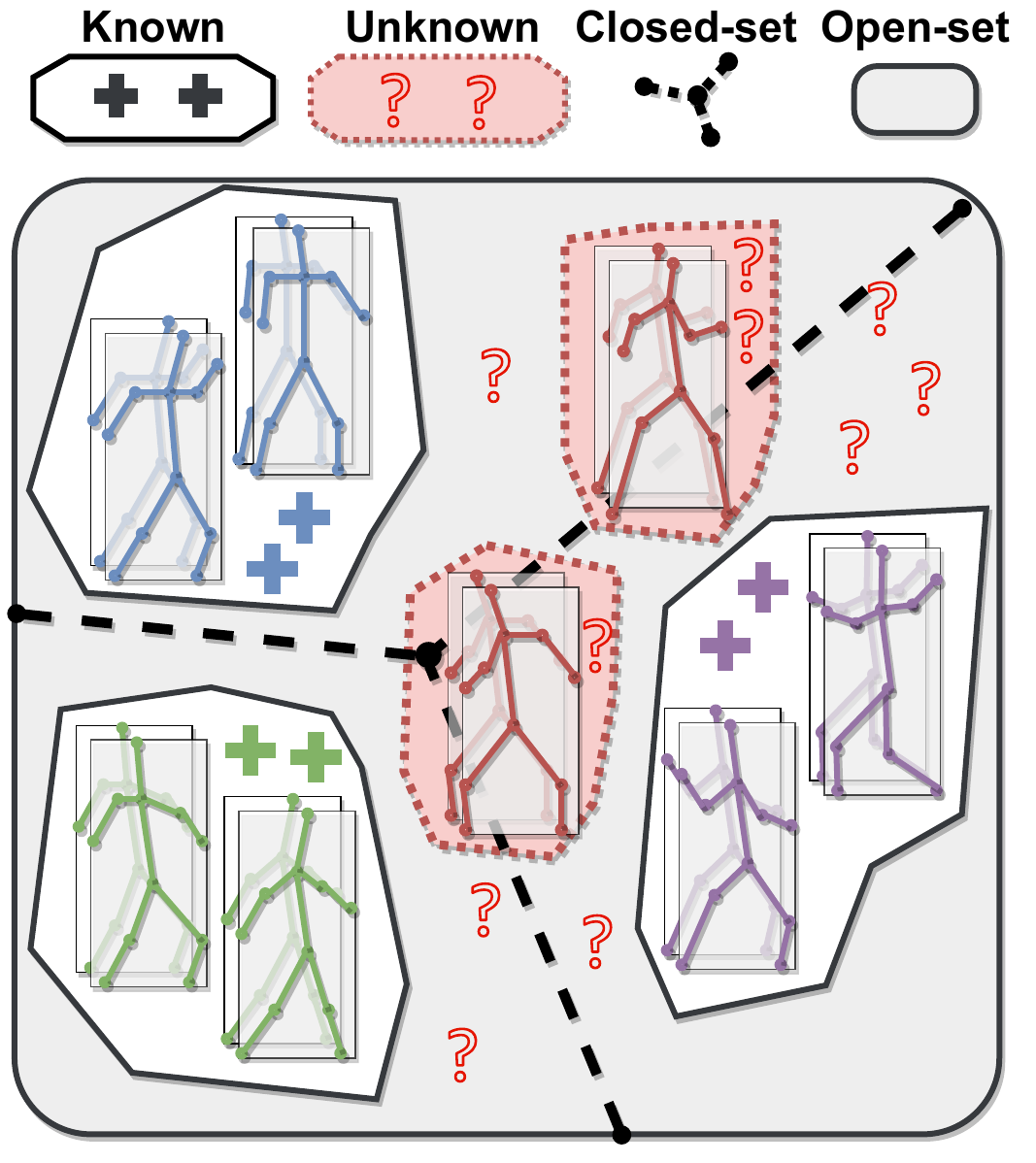}
        \centering
        \subcaption{Open-set setting}\label{fig1_a}
    \end{minipage}%
    \begin{minipage}[t]{0.55\columnwidth}
        \includegraphics[width=0.99\linewidth]{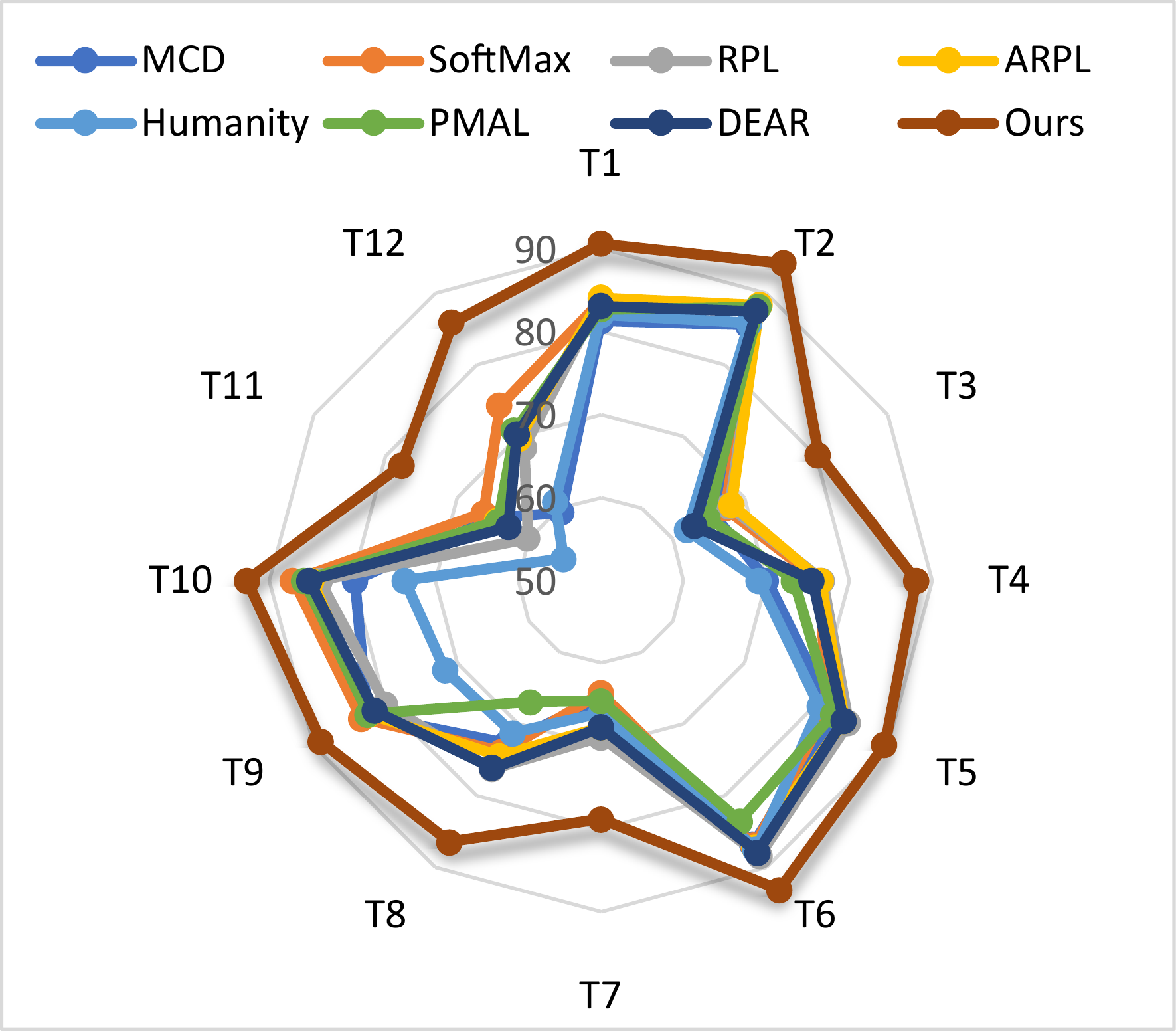}
        \centering
        \subcaption{Result comparison}\label{fig1_b}
    \end{minipage}%
    \caption{(a) The open-set skeleton-based action recognition setting. (b) Compared to previous methods, our method consistently achieves state-of-the-art performance. The tasks T1-T4 are based on the CTR-GCN backbone and use Cross-Subject and Cross-View splits of NTU60 to evaluate with O-AUROC and O-AUPR metrics, T5-T8 are with HD-GCN, and T9-T12 are with Hyperformer, respectively. }
    \label{fig:1}
\end{figure}

Leveraging body pose sequences for human action recognition offers several benefits, such as enhanced privacy, reduced data volume, and better generalization to novel human appearances. 
Modern skeleton-based approaches~\cite{zhou2022hypergraph}, once trained, remain static in their set of possible predictions.
A more realistic scenario is the model's exposure to \textit{open sets}, where both, known and novel action categories may occur at any time~\cite{meyer2019importance}. Out-of-distribution actions -- those that fall outside the model's known repertoire -- typically result in misclassifications as one of the known categories, eventually leading to significant disruptions, particularly when these recognition outputs directly steer decision-making, \eg, in assistive robots. As pointed out by researchers in the past~\cite{miller2018dropout, fontanel2020boosting} there is a pressing need for further exploration in open-set, skeleton-based human action recognition, which is the main motivation of our work.

Several methods target open-set action recognition \textit{in videos}~\cite{bao2021evidential}, but the problem of detecting novel behaviors from \textit{skeleton} streams has been overlooked so far.
These tasks pursue similar goals, yet differ substantially: the absence of visual background as an additional cue context and the sparse characteristic structure of body pose sequences introduce unique challenges in managing out-of-distribution actions. To address the lack of a suitable evaluation testbed, we first build an expansive benchmark for \textbf{O}pen-\textbf{S}et \textbf{S}keleton-based \textbf{A}ction \textbf{R}ecognition (OS-SAR), comprising three prominent skeleton-based action recognition backbones: CTRGCN~\cite{chen2021channel}, HDGCN~\cite{lee2022hierarchically}, and Hyperformer~\cite{ding2023hyperformer}. This benchmark is derived from three public datasets for action recognition from body pose sequences -- NTU60~\cite{shahroudy2016ntu}, NTU120~\cite{liu2020ntu}, and ToyotaSmartHome~\cite{dai2022toyota}) -- for which we formalize the open-set splits and an evaluation protocol. Effective and generalizable open-set recognition techniques should maintain stable performance for diverse combinations of datasets and backbones. 
Following the open-set recognition practices in image classification~\cite{lu2022pmal}, we randomly sample sets of unseen classes and compute the averaged performance over five random splits. 
However, presumably due to inherent differences between image/video and skeletal data, common open-set recognition strategies struggle to deliver consistent OS-SAR results and the recognition quality considerably fluctuates when considering different backbones and datasets.
This inconsistency underscores that the current methodologies struggle when deployed for OS-SAR challenges. 
A deeper examination reveals that the predicted open-set probability estimates of the existing methods are not realistic when exposed to a mix of in- and out-of-distribution skeletal sequences, which detrimentally affects open-set performance, steering models towards unwarranted overconfidence.

To tackle this problem, we introduce a new approach for OS-SAR.
Our method is multimodal and builds on three streams: joints, velocities, and bones, which we enable distribution-wise information exchange in their latent space via a novel Cross-Modality Mean Max Discrepancy (CrossMMD) suppression mechanism.
We also need to address the overconfidence of the SoftMax-normalized probability estimation when mixing in- and out-of-distribution samples~\cite{liu2020ntu}.
To this intent, we introduce a distance-based confidence measure based on the \textbf{C}hannel \textbf{N}ormalized \textbf{E}uclidean distance (CNE-distance) to the nearest latent space embeddings from the training set. 
This distance-based approach significantly improves the open-set recognition performance but falls short when it comes to the conventional close-set results compared to the vanilla SoftMax.
To have the best of both worlds, we propose a cross-modality distance-based logits refinement technique, which combines logits averaged across the modalities and the proposed CNE-distances.
We refer to our complete method as CrossMax, as it considers both, CrossMMD during training and cross-modality distance-based refinement during testing. CrossMax achieves state-of-the-art performances across datasets, backbones, and evaluation settings, shown in Fig.~\ref{fig:1}.

Our main contributions are as follows:
\begin{itemize}

  \item A large-scale benchmark for Open-Set Skeleton-based Action Recognition (OS-SAR), featuring three datasets for classification from body pose sequences, seven open-set recognition baselines, and three well-established backbones for skeleton data streams.

   \item  A multimodal approach for OS-SAR leveraging three streams: joints, velocities, and bones, and enabling the distribution-wise information exchange among them using the novel Cross-Modality Mean Max Discrepancy (CrossMMD) suppression mechanism.

 \item A distance-based confidence measure, the Channel Normalized Euclidean distance (CNE-distance), to address overconfidence in SoftMax-normalized probability estimates and enhance open-set recognition.

 \item The complete CrossMax methodology combines the aforementioned CrossMMD and the distance-based logits refinement technique, achieving state-of-the-art performance across various evaluations.

\end{itemize}

\section{Related Work}
\noindent\textbf{Skeleton-based action recognition} aims at recognizing action categories using the skeletal geometric information~\cite{ke2017new, liu2017enhanced, duan2022revisiting}. Most well-established methods are graph convolutional neural networks (GCN)-based~\cite{kipf2016semi, yan2018spatial, shi2019two, cheng2020decoupling, ye2020dynamic, chen2021channel}, more recent approaches leverage transformer architectures~\cite{shi2020decoupled, plizzari2021spatial, lee2022hierarchically, zhou2022hypergraph, ding2023hyperformer, xin2023transformer}. CTRGCN~\cite{chen2021channel}, HDGCN~\cite{liang2019hierarchical}, and Hyperformer~\cite{ding2023hyperformer} serve as backbones in our OS-SAR experiments due to their superior performances and large architecture discrepancy which allows for an evaluation regarding cross-backbone generalizability.

\noindent\textbf{Open-set recognition}, aiming at distinguishing classes, unseen during training~\cite{scheirer2012toward}, is nearly overlooked by the community for the task of skeleton-based action recognition, related works are mostly conducted in other fields, \eg, image classification and video-based action recognition. 
\cite{berti2022one} presented an approach for one-shot OS-SAR, but do not present methods for the general OS-SAR task. Due to the large discrepancy regarding this task, we resort to several well-established open-set image classification and open-set video-based action recognition approaches which can be adapted for OS-SAR by replacing backbone and input data. Shi \etal~\cite{shi2023open} proposed an OS-SAR approach using a 3D neural network on joints heat map as the backbone with deep evidential learning, which can be regarded as an implementation of DEAR~\cite{bao2021evidential}, while no comprehensive OS-SAR benchmark is contributed and the datasets leveraged are not commonly used in skeleton-based action recognition. We implement this approach by substituting the backbone into different GCNS in our benchmark since GCN is the dominant backbone to handle skeleton data.
\underline{In the field of open-set image classification}, multiple works~\cite{hendrycks2016baseline,yoshihashi2019classification,sun2020conditional,chen2020learning,chen2021adversarial,lu2022pmal, geng2020collective, oza2019c2ae} were presented. \cite{hendrycks2016baseline} first used the highest SoftMax score as the open-set probability, followed by reconstruction-based approaches~\cite{yoshihashi2019classification, oza2019c2ae, sun2020conditional, cen2023devil}. Recently, the most promising works are prototype-based methods~\cite{chen2020learning, chen2021adversarial, sun2020conditional}. Reciprocal points distance served as open-set probability in \cite{chen2020learning} and \cite{chen2021adversarial} while PMAL~\cite{lu2022pmal} is the state-of-the-art approach. \cite{cen2023devil} proposed a new task for unified few-shot open-set recognition. We choose to use SoftMax, RPL, ARPL, and PMAL as OS-SAR baselines. SoftMax could serve as a lower bound for OS-SAR while the rest have large potential to deliver superior performances in OS-SAR due to the success of these methods in the open-set image classification. 
\underline{In open-set video-based action recognition task}, at the early stage, \cite{shu2018odn} proposed Open Deep Network (ODN)
by adding novel classes incrementally to the recognition head to achieve awareness of new classes. \cite{krishnan2018bar} and \cite{subedar2019uncertainty} leveraged bayesian neural networks to achieve reliable uncertainty estimation. DEAR~\cite{bao2021evidential} constructed a large-scale benchmark for open-set video-based human action recognition. They also proposed an architecture that uses deep evidential learning and delivers state-of-the-art performance. Humpty Dumpty~\cite{du2023reconstructing} (renamed as Humpty in our benchmark) uses clip-wise relational graphical reconstruction error as the open-set probability.
Monte Carlo Dropout with Voting (MCD-V) is proposed by \cite{roitberg2020open} for open-set video-based driver action recognition. \cite{yang2019open} leveraged micro-doppler radar data, we do not adapt this model due to its specific architecture for such a modality.
DEAR, Humpty, and MCD-V serve as OS-SAR baselines. These baselines do not show consistent performances across datasets and backbones, displaying the need for a generalizable OS-SAR method. Thereby, we propose CrossMax which uses cross-modality mean max suppression in the training to enable cross-modality information exchange, and cross-modality distance-based logits refinement in the testing to refine the salient and non-salient logits position separately, introduced in the following section in detail.

\section{Method}
\begin{figure*}[t!]
\centering
\includegraphics[width=1\linewidth]{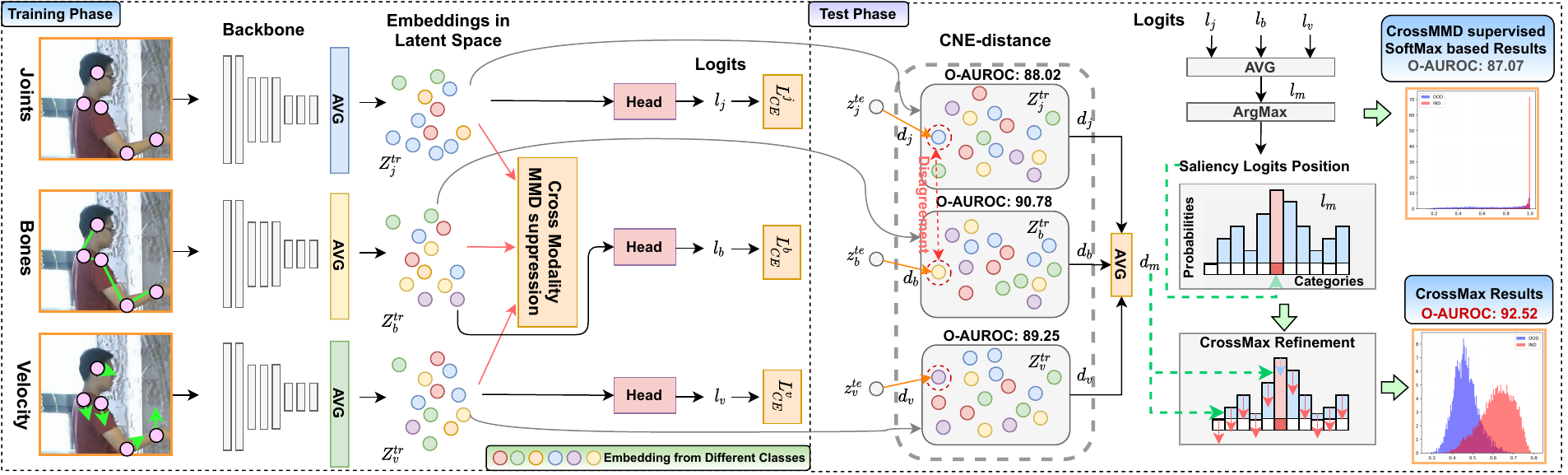}
\vskip-1.5ex
\caption{An overview of CrossMax. 
During training, we utilize the Cross-modality Mean Maximum Discrepancy (CrossMMD), to better align the latent spaces across different modalities. At test-time, for each modality, we calculate the Euclidean distance to the closest training set sample and combine this with the averaged logits from the three branches. This combination undergoes a refinement process based on the cross-modality distance, which is conducted differently on the salient and not-salient logits. The refined logits are then processed through SoftMax, a better confidence estimate for both in- and out-of-distribution samples, while keeping the accurate close-set classification capability inherent to the standard SoftMax.
}
\vskip-3ex
\label{fig:main}
\end{figure*}

\subsection{Benchmark}

We introduce OS-SAR, a large-scale benchmark for \textbf{O}pen-\textbf{S}et \textbf{S}keleton-based \textbf{A}ction \textbf{R}ecognition, leveraging CTR-GCN~\cite{chen2021channel}, HDGCN~\cite{liang2019hierarchical}, and Hyperformer~\cite{ding2023hyperformer} as the backbones to validate the generalizability of OS-SAR across different skeleton representations.
We build on the NTU60~\cite{shahroudy2016ntu}, NTU120~\cite{liu2020ntu}, and ToyotaSmartHome~\cite{dai2022toyota} datasets for human action recognition from body pose sequences and adapt their splits to suit open set conditions. Backbones and baselines are presented in the following, while the dataset introduction will be covered in the experiments section.
\subsubsection{Skeleton Representation Backbones.}
\noindent\underline{\textit{CTRGCN}}~\cite{chen2021channel} used Channel-wise Topology Refinement Graph Convolution (CTRGC) to dynamically learn distinct topologies and efficiently aggregate features in different channels within graph convolutional network (GCN). 
\noindent\underline{\textit{HDGCN}}~\cite{liang2019hierarchical} is based on Hierarchically Decomposed (HD) GCN by leveraging an HD-Graph that decomposes nodes into multiple sets to capture both structurally adjacent and distant edges with semantic relevance.  
\noindent\underline{\textit{Hyperformer}}~\cite{ding2023hyperformer} is a transformer-based approach that incorporates bone connectivity via graph distance embedding. 
We selected these architectures to validate the cross-backbone generalizability of open-set methods due to their strong performance in conventional skeleton-based human action recognition benchmarks and the different building blocks of their underlying architectures.

\subsubsection{Existing Open-Set Recognition Baselines.}
Our baselines are open-set recognition methods from the image classification task and the video-based action recognition task, as there are no methods specifically designed for open-set recognition from body poses yet, which can be adapted to diverse skeleton-based backbones. \noindent\underline{\textit{Open-set baselines from image classification:}} We selected principal-point distance-based approaches, \ie, RPL~\cite{chen2020learning} and ARPL~\cite{chen2021adversarial}, prototype learning-based approach, \ie, PMAL~\cite{lu2022pmal}, which is the current state-of-the-art approach in open-set image classification field, and the vanilla SoftMax score~\cite{hendrycks2016baseline} as baselines. \noindent\underline{\textit{Open-set baselines from video-based action recognition:}} We choose DEAR~\cite{bao2021evidential}, which uses deep evidential learning for open-set probability estimation, Monte Carlo Dropout + Voting (MCD-V)~\cite{roitberg2020open}, and Humpty~\cite{du2023reconstructing}, which uses temporal graph reconstruction as the open-set probability. We use the aforementioned skeleton backbones and skeleton data individually in all the selected open-set recognition baselines to achieve a fair comparison, where the image/video backbone and data are replaced by the selected skeleton-based backbones and skeleton data.

\subsection{CrossMax}

We propose CrossMax, a novel OS-SAR method leveraging three complementary skeleton modalities: joints, bones, and velocities. 
CrossMax first employs ensembled backbones for feature extraction while using Cross-modality Mean-Max Discrepancy suppression (CrossMMD) in training to enhance information exchange and reduce modality disparities. 
We further introduce a novel cross-modality distance-based logits refinement using Channel-Normalized Euclidean distance (CNE-distance). This refinement method significantly improves open-set probability estimation and close-set classification, as demonstrated in Fig.~\ref{fig:main}.

\subsubsection{Skeleton Modalities.}
 Given skeleton joints as $j =\{j_1^{\{1,..., N_j\}}, ..., j_T^{\{1,..., N_j\}}\}$, where $T$ denotes the total frame number of the skeleton sequence and $N_j$ indicates the joint number, bones $b$ and velocities $v$ can be calculated through $v =\{j_{t}^{\{1,..., N_j\}}{-}j_{t-1}^{\{1,..., N_j\}}\vert t \in \left[1, T\right]\}$, indicating the velocity of joints during motion at timestamp $t$, and $b =\{ j_{\{1,...,T\}}^x{-}j_{\{1,...,T\}}^y 
\vert (x,y)\in\Omega_{b}\}$, indicating the bone vector, where $\Omega_{b}$ indicates the set of bones, as depicted in Fig.~\ref{fig:main}.
\subsubsection{CrossMMD.}
To enable better cross-modal exchange, we present CrossMMD. 
The Mean Maximal Discrepancy (MMD) is used to quantify the dissimilarity between probability distributions~\cite{gretton2012kernel}. In our approach, MMD serves as a loss function, encouraging greater similarity between distributions. Note, that there is a lack of MMD-related research tailored for cross-modality scenarios.
Our primary goal is to diminish the significant discrepancy between latent spaces originating from diverse modalities. By doing so, we aim to leverage the intrinsic open-set discriminative cues to grasp the advantage of each branch, thereby facilitating an exchange of information based on distribution. We introduce the Cross-modality Mean Max Discrepancy Suppression Mechanism (CrossMMD) to address this challenge.
The Gaussian kernel is chosen in our approach to Reproduce Kernel Hilbert Space (RKHS). 
Let $\Omega_x$ and $\Omega_y$ denote two embedding batches, which can be interpreted as two distributions.  
First, we concatenate them to form $z = Concat(\Omega_x,\Omega_y)$. Then, we compute the pairwise L2 Norm distance between all samples within $z$ denoted as $d_z$. The bandwidths are then chosen according to Eq.~\ref{eq:bw} to determine the scales of the kernel function, which are influenced by both the sum of the distances and the sample number,
\begin{equation}
\label{eq:bw}
   BW=\frac{\sum(d_z)}{(N_z)^2 - N_z}, 
\end{equation}
where $N_z$ denotes the sample number. Let $N_k$ indicate the kernel number, we obtain the bandwidth list $L_{BW}$ as $\{BW* (\alpha)^i \vert ~i \in \left[0, N_k\right)\}$, where $\alpha$ is a scaling factor. Small bandwidths focus on capturing fine-grained dissimilarities among embeddings, which can be useful when the distributions have intricate local shapes. Large bandwidths capture broader cues and global discrepancies.
The kernel matrix of the given embeddings is obtained by,
\begin{equation}
\mathbb{H}_{k} = \{exp(-\frac{d_z}{\beta})\vert \beta \in L_{BW}\}.
\end{equation}
For $\mathcal{K} \in \mathbb{H}_k$, the intra-source differences can be calculated via Eq.~\ref{eq:intra}, where $\mathbb{E}$ indicates empirical mean average, $Z_{j}^{tr}$, $Z_{b}^{tr}$, and $Z_{v}^{tr}$ indicate the embeddings learned from three modalities during training as shown in Fig.~\ref{fig:main},
\begin{equation}
\label{eq:intra}
\begin{split}
    &\mathrm{Intra}(Z_{j}^{tr}, Z_{b}^{tr}, Z_{v}^{tr}) = \mathbb{E}\left[\sum_{\mathcal{K} \in \mathbb{H}_k}\mathcal{K}(Z_{j}^{tr}, Z_{j}^{tr})\right] + \\&\mathbb{E}\left[\sum_{\mathcal{K} \in \mathbb{H}_k}\mathcal{K}(Z_{b}^{tr}, Z_{b}^{tr})\right]
    + \mathbb{E}\left[\sum_{\mathcal{K} \in \mathbb{H}_k}\mathcal{K}(Z_{v}^{tr}, Z_{v}^{tr})\right],
\end{split}
\end{equation}
while the inter-source differences among different modalities can be calculated by Eq.~\ref{eq:inter},
\begin{equation}
\label{eq:inter}
\begin{split}
    &\mathrm{Inter}(Z_{j}^{tr}, Z_{b}^{tr}, Z_{v}^{tr}) = \mathbb{E}\left[\sum_{\mathcal{K} \in \mathbb{H}_k}\mathcal{K}(Z_{j}^{tr}, Z_{v}^{tr})\right] + \\&\mathbb{E}\left[\sum_{\mathcal{K} \in \mathbb{H}_k}\mathcal{K}(Z_{j}^{tr}, Z_{b}^{tr})\right]
    + \mathbb{E}\left[\sum_{\mathcal{K} \in \mathbb{H}_k}\mathcal{K}(Z_{b}^{tr}, Z_{v}^{tr})\right].
\end{split}
\end{equation}
The final CrossMMD is calculated as the discrepancy between the intra- and inter-source differences as Eq.~\ref{eq:mmd},
\begin{equation}
\label{eq:mmd}
\begin{split}
    &\mathrm{CrossMMD}(Z_{j}^{tr}, Z_{b}^{tr}, Z_{v}^{tr}) = \mathrm{Intra}(Z_{j}^{tr}, Z_{b}^{tr}, Z_{v}^{tr}) -\\ &\mathrm{Inter}(Z_{j}^{tr}, Z_{b}^{tr}, Z_{v}^{tr}).
    \end{split}
\end{equation}
CrossMMD is chosen as a loss function $L_{MMD}$ to enable the multi-scale information exchange among different modalities on comparable scales provided by the Gaussian kernels. 
Apart from the $L_{MMD}$, we also use the cross-entropy loss on the training set, depicted as Eq.~\ref{eq:loss},
\begin{equation}
\label{eq:loss}
    L_{\text{{overall}}} = L_{\text{{CE}}}^j +L_{\text{{CE}}}^b+ L_{\text{{CE}}}^v + \lambda \cdot L_{\text{{MMD}}},
\end{equation}
where $\lambda$ is chosen as a fixed value to keep the two losses having the same gradients scale. $ L_{\text{{CE}}}^j$,  $ L_{\text{{CE}}}^b$, and $ L_{\text{{CE}}}^v$ denote cross entropy losses for three branches, respectively.
\subsubsection{Cross-modality Distance-based Logits Refinement.}

By utilizing the averaged logits from three branches constrained by CrossMMD, the model yields a predicted open-set probability by using the highest score from SoftMax on the logits. However, though the performance of the open-set probability prediction increases, we find that SoftMax-based probability prediction suffers from a bad disentanglement between in- and out-of-distribution samples in terms of the open-set probability distribution, which limits the further improvement of OS-SAR.

To overcome this limitation, we propose a Channel-Normalized Euclidean distance (CNE-distance). This mechanism achieves Gaussian-wise probability distributions and ensures better disentanglement between the in- and out-of-distribution samples.
We first extract the embedding for the training samples considering three modalities and obtain the embedding sets as $Z_{a}^{tr}$, where $a\in \{j,b,v\}$. Then, we follow the same procedure to extract the embedding of the test sample, \ie, $z_{a}^{te}$. For each sample from the test set, we can obtain three distances according to the corresponding nearest embedding in the $Z_{a}^{tr}$, \ie, $d_{j}$, $d_{b}$, and $d_{v}$. We first utilize L2 normalization along the channel dimension for each embedding to map the feature value between $0$ and $1$. Then the Euclidean distance to the nearest training set embedding is used as the open-set probability. In summary, the CNE-distance can be calculated as Eq.~\ref{eq:distance},
\begin{equation}
\label{eq:distance}
\begin{split}
&d_{j},d_{b},d_{v}  = D\left[\mathcal{N}_C(z_{j}^{te}), \mathcal{N}_C(Z_{j}^{tr})\right],\\
 &D\left[\mathcal{N}_C(z_{b}^{te}), \mathcal{N}_C(Z_{b}^{tr})\right],D\left[\mathcal{N}_C(z_{v}^{te}), \mathcal{N}_C(Z_{v}^{tr})\right],
\end{split}
\end{equation}
where $te$ and $tr$ indicate the test and training set and $D\left[\cdot\right]$ is Euclidean distance. $\mathcal{N}_C (\cdot)$ indicates the channel normalization.
The averaged distance can be obtained by Eq.~\ref{eq:avg},
\begin{equation}
\label{eq:avg}
d_{m} = Mean(d_{j}, d_{b}, d_{v}).
\end{equation}
Our experiments reveal the effectiveness of the CNE-distance in producing more reliable probability estimates under open-set conditions, especially when differentiating between in- and out-of-distribution samples. 
Yet, when using the CNE-distance to determine the class among the known classes, as in Fig.~\ref{fig:main}, the results are sub-optimal. 
\begin{table*}[t]
\centering
\caption{Experiments on NTU60~\cite{shahroudy2016ntu}, NTU120~\cite{liu2020ntu}, and ToyotaSmartHome~\cite{dai2022toyota}, where CS, CV, and B indicate \textbf{C}ross-\textbf{S}ubject/\textbf{V}iew evaluations and \textbf{B}ackbone. The results are averaged for five random splits.}
\vskip-1ex
\label{tab:main}
\scalebox{0.6}{\begin{tabular}{l|l|llllll|llllll|llllll} 
\toprule\toprule
\multirow{3}{*}{\textbf{B}} & \multirow{3}{*}{\textbf{Method}} & \multicolumn{6}{c|}{\textbf{NTU60}} & \multicolumn{6}{c|}{\textbf{NTU120}} & \multicolumn{6}{c}{\textbf{Toyota Smart Home}} \\ 
\cline{3-20}
 &  & \multicolumn{2}{l}{\textbf{O-AUROC}} & \multicolumn{2}{l}{\textbf{O-AUPR}} & \multicolumn{2}{l|}{\textbf{C-ACC}} & \multicolumn{2}{l}{\textbf{O-AUROC}} & \multicolumn{2}{l}{\textbf{O-AUPR}} & \multicolumn{2}{l|}{\textbf{C-ACC}} & \multicolumn{2}{l}{\textbf{O-AUROC}} & \multicolumn{2}{l}{\textbf{O-AUPR}} & \multicolumn{2}{l}{\textbf{C-ACC}} \\
 &  & \textbf{CS} & \textbf{CV} & \textbf{CS} & \textbf{CV} & \textbf{CS} & \textbf{CV} & \textbf{CS} & \textbf{CV} & \textbf{CS} & \textbf{CV} & \textbf{CS} & \textbf{CV} & \textbf{CS} & \textbf{CV} & \textbf{CS} & \textbf{CV} & \textbf{CS} & \textbf{CV} \\ 
\midrule
\multirow{8}{*}{\rotatebox[origin=c]{90}{CTRGCN}}%
 & SoftMax~\cite{hendrycks2016baseline} & 83.68 & 87.77 & 67.37 & 76.38 & 90.56 & 93.83& 82.37 & 83.10 & 91.84 & 91.88 & 90.37 & 91.04 & 70.04 & 65.18 & 70.10 & 69.02 & 70.41 &  78.52  \\
 & RPL~\cite{chen2020learning} & 84.02 & 88.06 & 67.86 & 76.75 & 90.82 & 95.38 & 82.06 & 83.40 & 91.55 & 92.05 & 90.40 & 90.96 & 56.74 & 51.90 & 60.46 & 59.46 & 74.42 &  75.41\\
 & ARPL~\cite{chen2021adversarial} & 84.13 & 88.37 & 68.24 & 76.58 & 91.00 & 95.45 & 81.93 & 83.03 & 91.54 & 91.80 & 90.12 & 91.16 & 74.11 & 64.22 & 73.80 & 67.04 & 78.55 & 79.53 \\
 & PMAL~\cite{lu2022pmal} & 82.72 & 88.06 & 64.99 & 73.31 & 90.74 & 95.09 & 80.46 & 81.75 & 90.55 & 90.93 & 89.61 & 90.14 & 57.80 & 51.73 & 61.27 & 52.94 & 74.06 & 67.50 \\
 & DEAR~\cite{bao2021evidential} & 83.11 & 87.54 & 63.07 & 75.52 & 84.14 & 95.41 & 81.98 & 82.66 & 91.51 & 91.67 & 90.11 & 90.61 & 76.19 & 60.54 & 75.42 & 74.52 & 78.49 & 65.50 \\
& Humpty~\cite{du2023reconstructing} & 82.08 & 85.82 & 62.05 & 69.09 & 89.17 & 93.75 & 82.12 & 83.35 & 90.78 & 91.06 & 89.89 & 90.54 & 65.10 & 59.17 & 68.71 & 62.43 & 77.76 & 75.19 \\
 & MCD-V~\cite{roitberg2020open} & 81.31 & 85.58 & 61.88 & 69.99 & 90.14 & 94.72 & 78.83 & 79.17 & 89.60 & 76.93 & 88.12 & 88.10 & 69.61 & 67.92 &71.12  & 71.68 & 77.74 & 76.41 \\
 \cline{2-20}
  &{\cellcolor{LightCyan}}Ours &{\cellcolor{LightCyan}}\textbf{90.62} &{\cellcolor{LightCyan}}\textbf{94.14} &{\cellcolor{LightCyan}}\textbf{80.32} &{\cellcolor{LightCyan}}\textbf{88.07} & {\cellcolor{LightCyan}}\textbf{93.68} & {\cellcolor{LightCyan}}\textbf{97.51} &{\cellcolor{LightCyan}}\textbf{85.44} &{\cellcolor{LightCyan}}\textbf{85.42} & {\cellcolor{LightCyan}}\textbf{93.67} &{\cellcolor{LightCyan}}\textbf{93.36} &{\cellcolor{LightCyan}}\textbf{91.43} & {\cellcolor{LightCyan}}\textbf{92.94} &{\cellcolor{LightCyan}}\textbf{83.99} &{\cellcolor{LightCyan}}\textbf{84.00} & {\cellcolor{LightCyan}}\textbf{86.74} &{\cellcolor{LightCyan}}\textbf{87.37} &{\cellcolor{LightCyan}}\textbf{80.25} & {\cellcolor{LightCyan}}\textbf{80.51} \\ 
 \hline
\multirow{8}{*}{\rotatebox[origin=c]{90}{HDGCN}}%
 & SoftMax~\cite{hendrycks2016baseline} & 81.52 & 86.95 & 63.62 & 73.89 & 89.14 & 94.67 & 81.34 & 82.90 & 91.49 & 91.83 & 89.92 & 90.21 & 72.88 & 54.47 & 71.16 & 61.07 & 78.37 & 75.10 \\
 & RPL~\cite{chen2020learning} & 82.92 & 88.38 & 66.06 & 76.27 & 91.92 & 95.32 & 82.00 & 83.05 & 91.59 & 91.83 & 89.77 & 90.77 & 74.26 & 61.93 & 73.97 & 63.61 & 78.35 & 77.02 \\
 & ARPL~\cite{chen2021adversarial} &  83.92  &  87.19 & 67.76 & 74.49 & 90.65 & 94.90&82.06 & 82.80 & 91.51 & 91.74 & 90.08 & 90.68 & 73.00 & 64.53 & 72.93& 68.73 & 78.75  & 77.62 \\
 & PMAL~\cite{lu2022pmal} & 82.41 & 83.57 & 64.53 & 66.98 & 90.26 & 93.33 & 80.68 & 81.89 & 90.71 & 91.22 & 89.53 & 90.75 & 64.64 & 74.72 & 69.20 & 73.41 & 77.23 & 78.39 \\
 & DEAR~\cite{bao2021evidential} & 83.87 & 87.92 & 67.76 & 76.15 & 90.65 & 95.15 & 81.89 & 82.78 & 91.38 & 91.63 & 89.85 & 90.68 & 75.03 &59.25  & 75.10 & 63.41 & 78.41 & 78.54 \\
 & Humpty~\cite{du2023reconstructing} & 81.91 & 87.47 & 61.49 & 71.32 & 88.70 & 94.64 & 82.38 & 83.26 & 90.72 & 85.78 & 89.40 & 89.93 & 62.41 & 57.12 & 66.78 & 66.32 & 77.83 & 80.12 \\
  & MCD-V~\cite{roitberg2020open} & 82.51 & 86.74 & 64.24 & 72.70 & 90.04 & 94.88 & 80.55 & 80.24 & 90.26 & 90.27 & 89.80 & 89.00 & 72.29 & 64.64 & 72.57 & 69.20 & 79.90 & 78.93 \\
 \cline{2-20}
  & {\cellcolor{LightCyan}}Ours &{\cellcolor{LightCyan}}\textbf{89.57} &{\cellcolor{LightCyan}}\textbf{93.14} & {\cellcolor{LightCyan}}\textbf{78.82} &{\cellcolor{LightCyan}}\textbf{86.48} &{\cellcolor{LightCyan}}\textbf{93.30} &{\cellcolor{LightCyan}}\textbf{96.88} &{\cellcolor{LightCyan}}\textbf{83.76} &{\cellcolor{LightCyan}}\textbf{84.46} & {\cellcolor{LightCyan}}\textbf{92.84} &{\cellcolor{LightCyan}}\textbf{93.07} &{\cellcolor{LightCyan}}\textbf{90.82} & {\cellcolor{LightCyan}}\textbf{91.67} & {\cellcolor{LightCyan}}\textbf{84.32}&{\cellcolor{LightCyan}}\textbf{83.70} &{\cellcolor{LightCyan}}\textbf{86.57} &{\cellcolor{LightCyan}}\textbf{86.44} & {\cellcolor{LightCyan}}\textbf{80.41} &{\cellcolor{LightCyan}}\textbf{81.29}  \\ 
 \hline
\multirow{8}{*}{\rotatebox[origin=c]{90}{HyperFormer}}%
 & SoftMax~\cite{hendrycks2016baseline} &  83.40& 87.11 & 66.29 & 74.38 & 90.46 & 94.90 & 81.16 & 82.74 & 91.40 & 91.60 & 90.69 & 90.95 & 74.25 & 72.26 & 74.30 & 74.94 & 78.68 & 81.40 \\
 & RPL~\cite{chen2020learning} & 79.97 & 83.96 & 60.15 & 68.52 & 88.39 & 92.46 & 81.26 & 82.20 & 91.19 & 91.30 & 89.65 & 90.31 & 73.24 & 74.30 & 72.62 & 75.84 & 78.62 & 82.23 \\
 & ARPL~\cite{chen2021adversarial} & 82.37 & 84.88 & 64.38 & 69.74 & 89.87 & 93.99 & 82.08 & 82.06 & 91.25 & 91.53 & 90.19 & 90.46 & 72.73 & 72.77 & 72.99 & 73.98 & 78.60 & 82.67 \\
 & PMAL~\cite{lu2022pmal} & 82.43 & 85.80 & 64.29 & 70.89 & 90.33 & 94.79 & 81.95 & 81.90 & 91.63 & 89.13 & 90.65 & 90.42 &  73.48& 51.89 & 73.68 & 47.26 & 78.01 & 69.97 \\
 & DEAR~\cite{bao2021evidential} & 81.47 & 85.22 & 62.87 & 70.33 & 89.94 & 94.26 & 81.00 & 81.90 & 90.96 & 91.15 & 89.51 & 90.15 & 72.86 & 74.54 & 72.70 & 76.09 &78.20  & 82.87 \\
 & Humpty~\cite{du2023reconstructing} & 71.72 & 73.66 & 55.21 & 60.95 & 89.98 & 94.67 & 70.67 & 69.28 & 86.93 & 86.29 & 89.92 & 89.40 & 72.32 & 62.70  & 71.26 &  62.88& 78.32 & 80.23 \\
  & MCD-V~\cite{roitberg2020open} & 82.52 & 79.69 & 65.00 & 59.46 & 93.05 & 88.80 & 80.21 & 81.17 & 90.24 & 90.64 & 88.87 & 89.80 & 61.69 & 53.71 & 65.17 & 62.70 & 74.15 & 48.20 \\
 \cline{2-20}

  & {\cellcolor{LightCyan}}Ours &{\cellcolor{LightCyan}}\textbf{88.98} & {\cellcolor{LightCyan}}\textbf{92.73} &{\cellcolor{LightCyan}}\textbf{77.75} &{\cellcolor{LightCyan}}\textbf{85.94} &{\cellcolor{LightCyan}}\textbf{93.24} &{\cellcolor{LightCyan}}\textbf{96.71} &{\cellcolor{LightCyan}}\textbf{83.67} &{\cellcolor{LightCyan}}\textbf{83.70} &{\cellcolor{LightCyan}}\textbf{92.84} & {\cellcolor{LightCyan}}\textbf{92.62} &{\cellcolor{LightCyan}}\textbf{91.30} &{\cellcolor{LightCyan}}\textbf{92.50} &{\cellcolor{LightCyan}}\textbf{82.23} & {\cellcolor{LightCyan}}\textbf{80.76} &{\cellcolor{LightCyan}}\textbf{84.28} & {\cellcolor{LightCyan}}\textbf{81.46} &{\cellcolor{LightCyan}}\textbf{79.58}  &{\cellcolor{LightCyan}}\textbf{83.54} \\
\hline
\hline
\end{tabular}}
\vskip-3ex
\end{table*}

To address this, we introduce a novel refinement methodology. This approach refines the averaged logits utilizing the CNE-distance, addressing the disparities among modalities and improving the close-set classification.
By incorporating the averaged CNE-distances among modalities, our method seeks to strike a balance between effective open-set probability estimation and good closed-set classification. 
We first acquire the position with the highest logit value of the averaged logits by Eq.~\ref{ea:avg_l},
\begin{equation}
\label{ea:avg_l}
M_P= ArgMax((l_{j} + l_{b} + l_{v})/3),
\end{equation}
where $l_j$, $l_b$, and $l_v$ denote the predicted logits for joints, bones, and velocities branches through classification heads.
Then we refine the predicted averaged logits $l_{m}$ by using Eq.~\ref{eq:refine} considering a given sample, where the salient logit position is indicated by a one-hot mask $M_P$,
\begin{equation}
\label{eq:refine}
    l_{m}\left[M_{P}\right] := Log((exp(l_{m}\left[M_{P}\right]*d_{m}^2))(\frac{1}{d_{m}}-1)).
\end{equation}
While the not salient positions are indicated by mask $M_{NP}$, the not saliency logits are refined by Eq.~\ref{eq:not_salient},
\begin{equation}
\label{eq:not_salient}
    l_{m}\left[M_{NP}\right] := l_{m}\left[M_{NP}\right]*d_{m}^2.
\end{equation}
Then, we get the refined full logits $l_{m}$, which will be passed through SoftMax further to get the classification and the open-set probability. The final predicted open-set probability is $P_{prob}{=}Max(SoftMax(l_{m}))$, while the open-set novelty score can be obtained by $1-P_{prob}$. By using this refinement method, the accurately predicted class from the SoftMax score computed on averaged logits can be preserved while the predicted open-set probability can achieve a distance-controllable disentanglement. This disentanglement ability benefits the OS-SAR a lot, as observed in our experiments. We refer to our full pipeline combining CrossMMD during training and the proposed distance-based refinement at test-time as CrossMax. CrossMax shows superior OS-SAR performances across all the backbones, and datasets, which will be discussed in detail in the next section. 

\section{Experiments}

\begin{table}
\centering
\caption{Module ablation on NTU60 cross-subject evaluation on CTRGCN backbone, where the results are averaged among five random splits.}
\vskip-1ex
\label{tab:ablation_module}
\scalebox{0.85}{\begin{tabular}{l|lll} 
\toprule
\toprule
\textbf{Method} & \textbf{O-AUROC} & \textbf{O-AUPR} & \textbf{C-ACC} \\ 
\midrule
Ensemble & 86.23 & 71.35 & 93.31 \\
\midrule
CrossMMD (Ours) & 88.31 & 74.80 & 93.68 \\
CrossMax (Ours) & 90.62 & 80.32 & 93.68  \\
\hline
\hline
\end{tabular}}
\vskip-3ex
\end{table}

\subsection{Datasets and Metrics}
\subsubsection{Datasets.}
\underline{NTU60}~\cite{shahroudy2016ntu} involves $56,880$ samples of $60$ action classes. We randomly choose $20$ classes as out-of-distribution classes.
\underline{NTU120}~\cite{liu2020ntu} involves $120$ action classes. We randomly choose $90$ classes as out-of-distribution classes.
\underline{ToyotaSmartHome}~\cite{dai2022toyota} contains $16,115$ samples with $31$ classes, which is challenging since occlusion from real-world scenarios is involved. $18$ action classes are selected as out-of-distribution classes. 
\subsubsection{Metrics.} The area under the receiver operating characteristic (\underline{O-AUROC}) and area under the precision-recall curve (\underline{O-AUPR}) are the most important metrics to evaluate the open-set performance with different focuses regarding the category balancing. Alongside, close-set classification accuracy (\underline{C-ACC}) is chosen as a minor indicator of whether the open-set method can preserve good classification capability or not. O-AUROC and C-ACC metrics are selected following PMAL~\cite{lu2022pmal}, while O-AUPR is additionally provided since ToyotaSmartHome is unbalanced. More details are delivered in the supplementary.

\subsection{Implementation Details}
Our method relies on PyTorch1.8.0 and is trained with SGD optimizer with learning rate (lr) $0.1$, step-wise lr scheduler with decay rate $0.1$, steps for decay at $\{35,55,70\}$, weight decay $0.0004$, and batch size $64$ for $100$ epochs on $4$ Nvidia A100 GPUs with Intel Xeon Gold 6230 processor. $\lambda$, $N_k$, and $\alpha$ are chosen as $0.1$, $5$, and $2.0$, respectively. In total, our method has $4.29$ MB, $5.04$ MB, and $7.8$ MB number of parameters on CTRGCN, HDGCN, and Hyperformer.

\subsection{Benchmark Analysis}
We first give a comprehensive analysis of the performances for existing open-set recognition approaches on the OS-SAR benchmark in Tab.~\ref{tab:main}. Taking cross-backbone generalizability into consideration, principal points distance-based approaches, \ie, RPL~\cite{chen2020learning} and ARPL~\cite{chen2021adversarial}, achieve $0.34\%$ and $0.45\%$ O-AUROC improvements on CTRGCN and $1.40\%$ and $2.40\%$ O-AUROC improvements on HDGCN compared with SoftMax~\cite{hendrycks2016baseline} on NTU60 for cross-subject evaluation. However, their performances are below SoftMax on the HyperFormer on NTU60, indicating that principal points distance-based approaches can not well generalize to different skeleton-based action recognition backbones.

Then we turn our concentration on the generalizability among different datasets. On NTU120, RPL and ARPL can achieve better performances compared with SoftMax on HDGCN and Hyperformer backbones, while on ToyotaSmartHome cross-subject evaluation, RPL fails to work well on CTRGCN and both of these two approaches only work better compared with SoftMax on HDGCN backbone. Considering the prototypical learning approach, \ie, PMAL~\cite{lu2022pmal}, it generally does not work well on the OS-SAR task. 
Three open-set approaches from video-based action recognition task, the deep evidential learning approach, \ie, DEAR~\cite{bao2021evidential}, the Monte Carlo Dropout + Voting approach, \ie, MCD-V~\cite{roitberg2020open}, and the temporal relationship reconstruction approach, \ie, Humpty~\cite{du2023reconstructing}, unfortunately, deliver limited performances for OS-SAR. This consequence is caused by the large differences between RGB image/video data and the skeleton data, as skeleton data lacks a majority of the background cues and visual appearance cues while the data format is quite sparse.
Another unignorable reason is that the networks utilized for feature extraction of skeleton data are mostly GCNs instead of convolutional neural networks (CNNs) or rely on graph architecture, where different manifolds on the latent space could be delivered due to the backbone discrepancy. This observation indicates the critical need to develop a generalizable OS-SAR approach that can work well across datasets and backbones.
\begin{figure}[t!]
\centering
\includegraphics[width=1\linewidth]{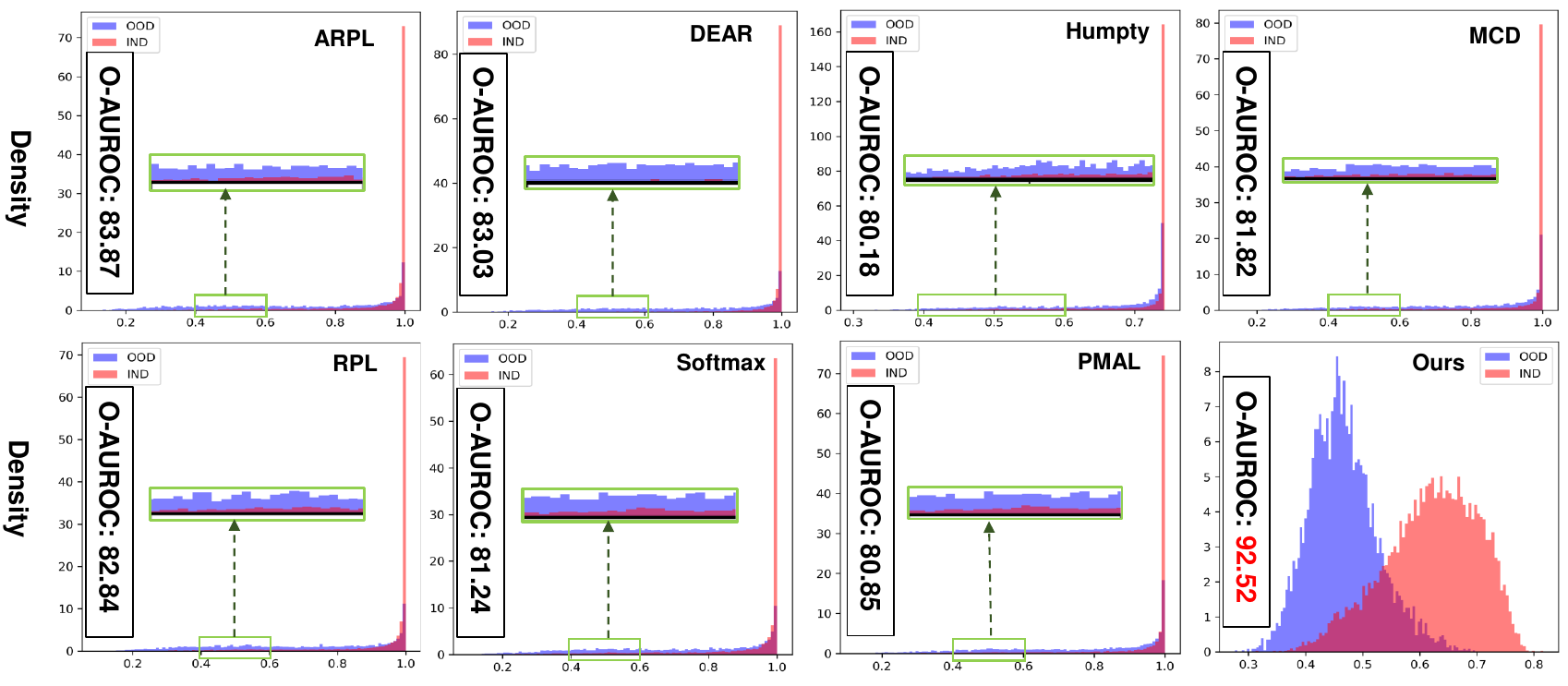}
\vskip-1ex
\caption{Comparison of the open-set probabilities using HD-GCN as the backbone on NTU60 for cross-subject evaluation for one random split. Our approach achieves better disentanglement of the in- and out-of-distribution samples.}
\vskip-3ex
\label{fig:probability_score1}
\end{figure}
To handle the underlying issue for existing open-set recognition approaches, we analyze the disentanglement between the in- and out-of-distribution samples considering the open-set probability in Fig.~\ref{fig:probability_score1}, where open-set probability tends to $1.0$ when the prediction is quite certain. We observe that most of the baselines can not well disentangle in- and out-of-distribution samples according to their predicted open-set probabilities, which serves as a critical reason for the undesired performance on OS-SAR. Keeping this issue in mind, we propose CrossMax by using CrossMMD in the training phase and cross-modality distance-based logits refinement in the test phase. CrossMax delivers superior disentanglement in terms of the open-set probability considering the in- and out-of-distribution samples. CrossMax achieves $6.94\%$, $8.05\%$, and $5.58\%$ O-AUROC improvements and $12.95\%$, $15.20\%$, and $11.46\%$ O-AUPR improvements on CTRGCN, HDGCN, and Hyperformer backbones within NTU60 cross-subject evaluation compared with vanilla SoftMax, while consistent performances can be found for different backbones, datasets, and settings, demonstrating the importance of the superior disentanglement ability for open-set probability between in- and out-of-distribution samples.
\subsection{Analysis of Observations and Ablations}
\subsubsection{Benefits by using CrossMMD.} 
\begin{figure}[t!]
\centering
\includegraphics[width=1\linewidth]{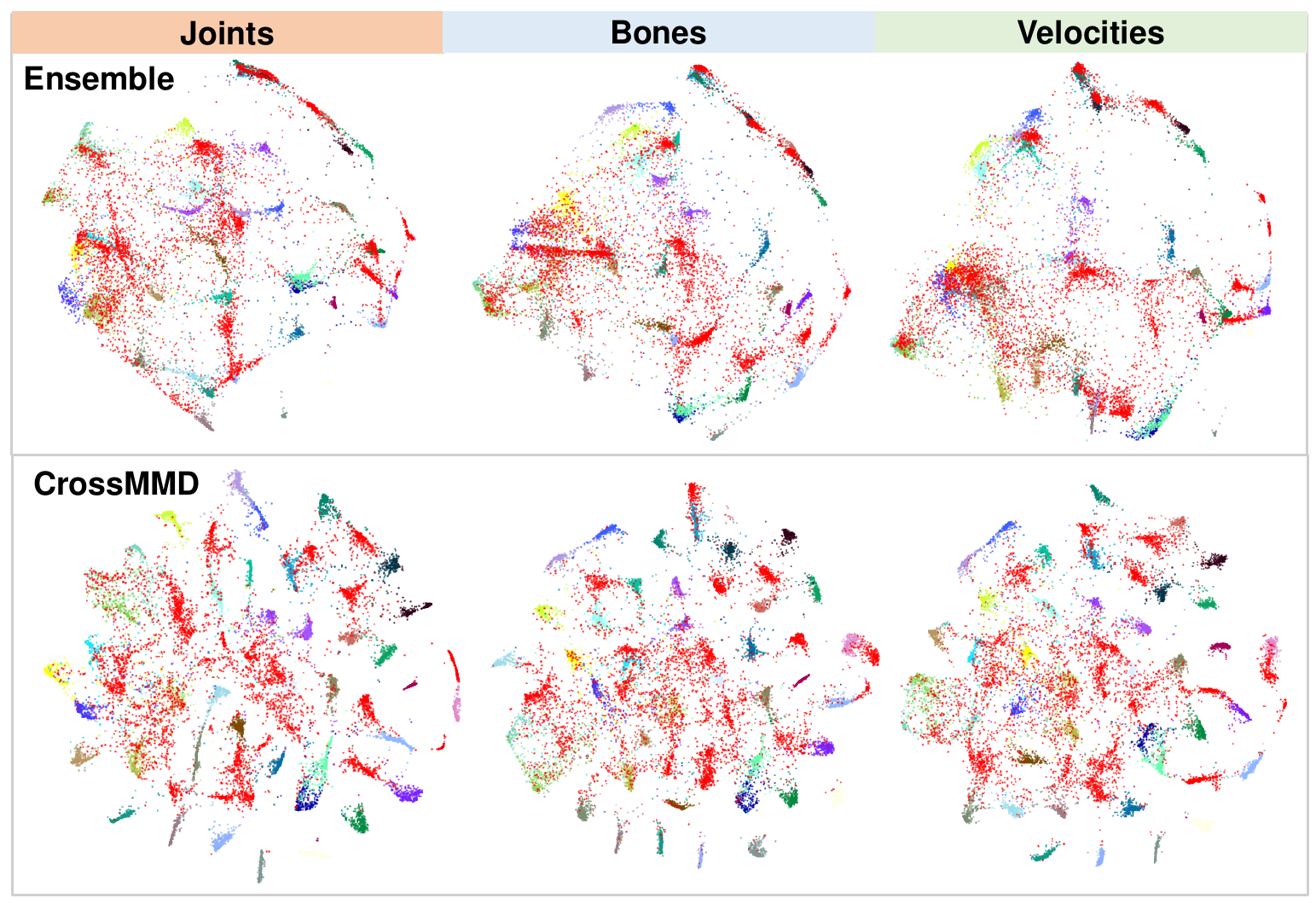}
\vskip-2ex
\caption{Comparison for t-SNE~\cite{JMLR:v9:vandermaaten08a} visualizations on NTU60 cross-subject evaluation with CTRGCN backbone. Out- and in-of-distribution samples are marked by red and other colors, respectively.}

\label{fig:TSNE}
\end{figure}

To analyze the benefits delivered by the CrossMMD, t-SNE visualizations are provided to illustrate the changes before using CrossMMD (marked as Ensemble), and after using CrossMMD (marked as CrossMMD), in Fig.~\ref{fig:TSNE}. We observe that by using CrossMMD, the latent spaces are more discriminative and structured for in- and out-of-distribution samples on all modalities, which matches the performance benefits introduced in Tab.~\ref{tab:ablation_module}, where Ensemble and CrossMMD both use vanilla SoftMax score to get the open-set probability estimation.

 \begin{figure}[t!]
\centering
\vskip -2ex
\includegraphics[width=1\linewidth]{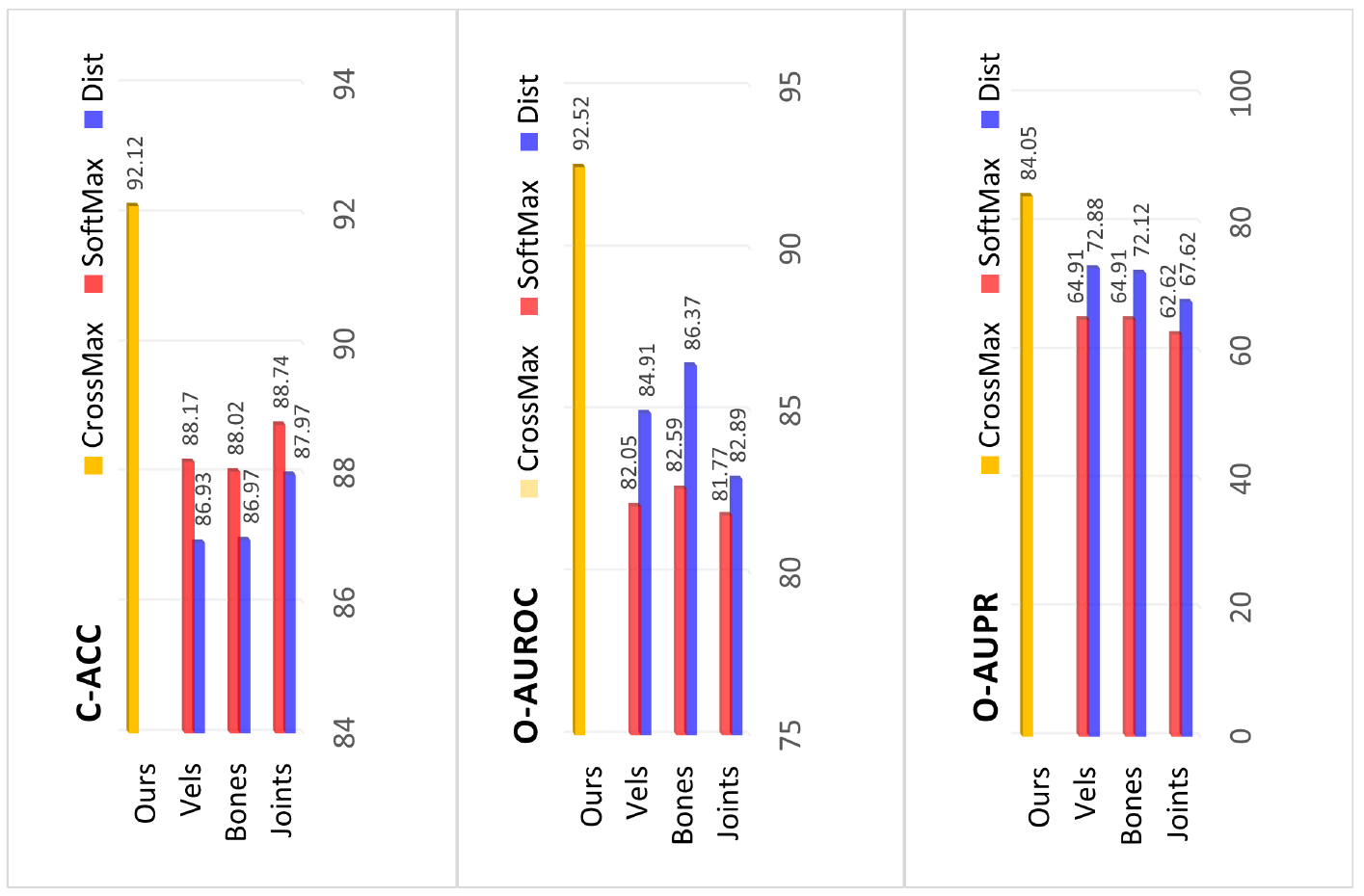}
\vskip-2ex
\caption{Performance comparison among SoftMax, CNE-distance, and CrossMax using HDGCN as the backbone on NTU60 cross-subject evaluation for one random split.} 
\vskip-3ex
\label{fig:Dist_VS_SoftMax}
\end{figure}
\begin{figure}[t]
    \centering
    \begin{minipage}[t]{0.5\columnwidth}
        \includegraphics[width=0.99\linewidth]{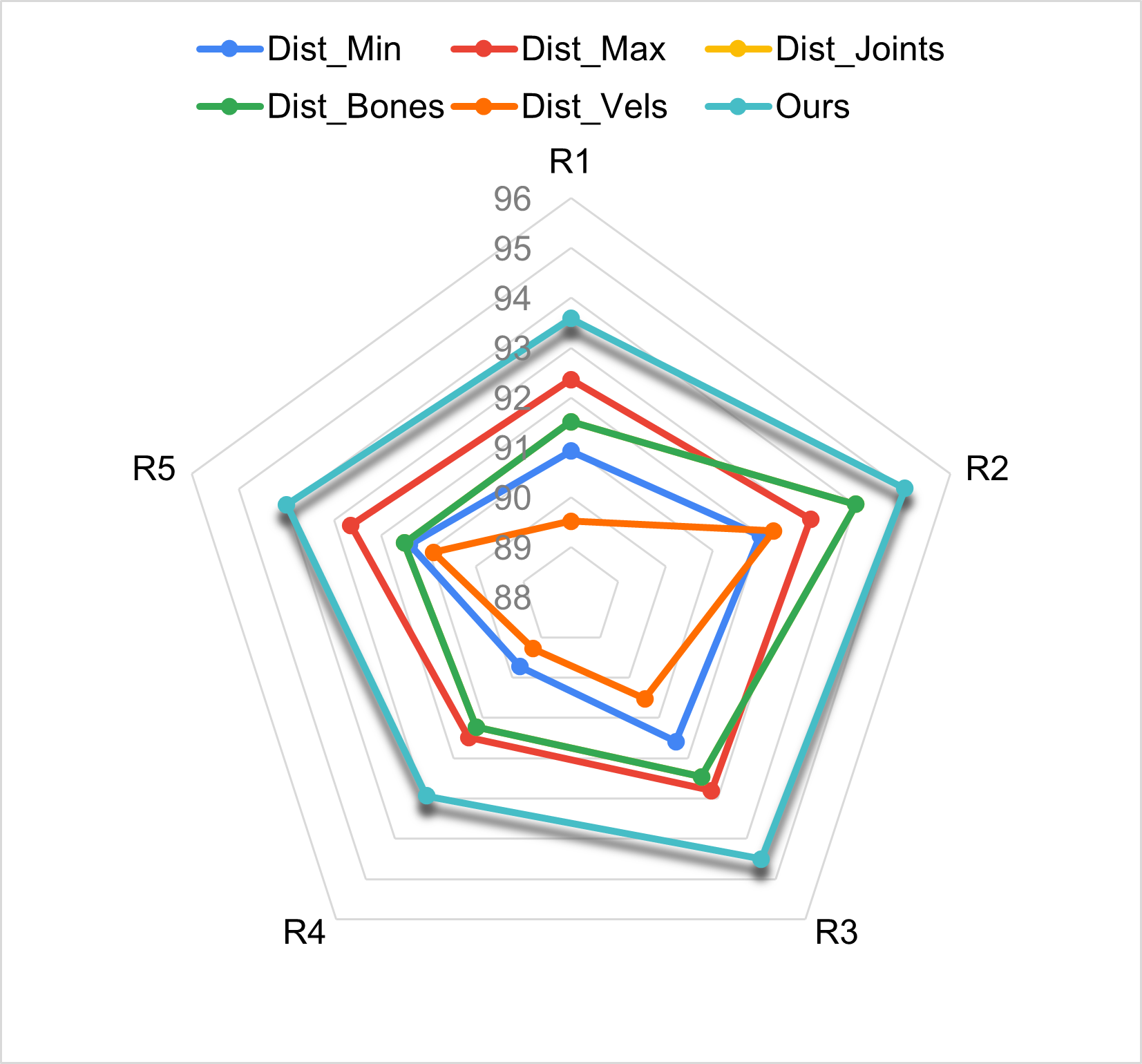}
        \centering
        \subcaption{O-AUROC}\label{score_a}
    \end{minipage}%
    \begin{minipage}[t]{0.5\columnwidth}
    \includegraphics[width=0.99\linewidth]{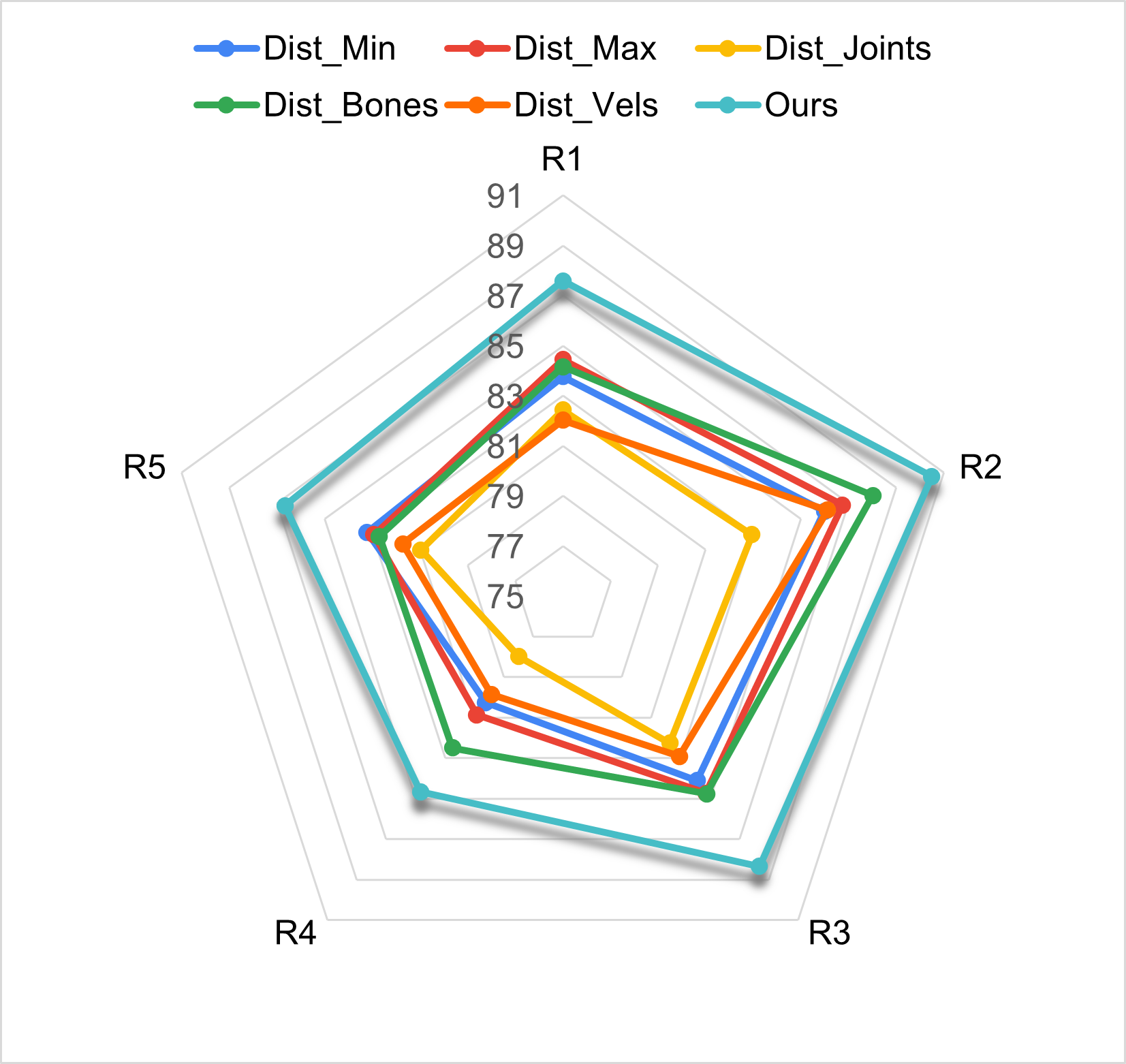}
        \centering
        \subcaption{O-AUPR}\label{score_b}
    \end{minipage}%
    \vskip-1ex
    \caption{Comparison of open-set recognition performances using CTRGCN backbone on NTU60 cross-view evaluation for five different random splits. }
    \label{fig:openset_score_dist_ablation}
\vskip-2ex
\end{figure}

\subsubsection{Comparison between CNE-distance \vs. vanilla SoftMax.} As mentioned in the introduction, we observe that the open-set recognition performances for O-AUROC and O-AUPR of CNE-distance on different modalities are much better compared with those of the vanilla SoftMax without our distance-based logits refinement and deliver the proof in Fig.~\ref{fig:Dist_VS_SoftMax}. Compared with vanilla SoftMax, CNE-distance can achieve $2.86\%$, $3.78\%$, and $1.12\%$ O-AUROC improvements for joints, bones, and velocities, respectively, while consistent results can be found for O-AUPR. However, as mentioned in the introduction, CNE-distance has a shortcoming in achieving a satisfied decision on the close-set classification, where the performances are worse compared with the vanilla SoftMax. This novel logits refinement approach takes advantage of both vanilla SoftMax and the CNE-distance and it achieves superior open-set probability estimation and close-set classification performances. 
\subsubsection{Comparison between logits refinement \vs. CNE-distance.}
 After the above-mentioned analysis, there would be a question regarding how well the proposed cross-modality distance-based logits refinement outperforms CNE-distance ablations in terms of the open-set probability prediction ability. To delve deeper into this question, we showcase a comparison in Fig.~\ref{fig:openset_score_dist_ablation} where the results are from five random splits marked as R1 to R5.
 We choose CNE-distance ablations as joint-modality distance (Dist\_joints), bone-modality distance (Dist\_bones), velocity-modality distance ((Dist\_velocities)), the min aggregation (Dist\_min), and the max aggregation (Dist\_max) over three branches. If the shape of the curve for one approach is a regular pentagon, the performance is stable and robust over different splits. We observe that the prediction by using our logits refinement achieves the best stable performance compared with the others by a large margin, indicating that using the cross-modality logits refinement method can even harvest a better open-set probability while preserving the superior close-set classification ability from SoftMax.

 \subsubsection{Ablation of each module.} We use CrossMMD during the training while using cross-modality distance-based logits refinement in the test phase. We showcase the benefits from different modules in Tab.~\ref{tab:ablation_module}, where \textit{Ensemble} indicates using three ensemble modalities during training while using vanilla SoftMax during testing, \textit{CrossMMD} indicates using CrossMMD during training while using vanilla SoftMax during testing, and \textit{CrossMax} indicates using CrossMMD during training while using cross-modality logits refinement during testing. \textit{CrossMMD} achieves $2.08\%$, $3.45\%$, and $0.37\%$ improvements for O-AUROC, O-AUPR, and C-ACC, while \textit{CrossMax} preserves the superior C-ACC of \textit{CrossMMD} and delivers large improvements by $2.31\%$ and $5.52\%$ of O-AUROC and O-AUPR, indicating the importance by using both of them for OS-SAR.

\section{Conclusion}
We propose the OS-SAR benchmark to contribute a large-scale test bed for open-set skeleton-based action recognition across backbones and datasets while selecting seven well-established open-set recognition methods serving as baselines.
We identify that most existing open-set recognition methods do not work well on OS-SAR and thereby propose CrossMax using CrossMMD during the training phase and cross-modality distance-based logits refinement during the test phase.
Our approach achieves state-of-the-art performances on OS-SAR while indicating great ability in disentangling the in- and out-of-distribution samples in terms of the predicted open-set probability.

\bibliography{bibfiles/other,bibfiles/skeleton_based_action_recognition, bibfiles/open_set_recognition, bibfiles/datasets,bibfiles/architectures}

\begin{thebibliography}{43}
\providecommand{\natexlab}[1]{#1}

\bibitem[{Bao, Yu, and Kong(2021)}]{bao2021evidential}
Bao, W.; Yu, Q.; and Kong, Y. 2021.
\newblock Evidential deep learning for open set action recognition.
\newblock In \emph{ICCV}.

\bibitem[{Berti et~al.(2022)Berti, Rosasco, Colledanchise, and Natale}]{berti2022one}
Berti, S.; Rosasco, A.; Colledanchise, M.; and Natale, L. 2022.
\newblock One-shot open-set skeleton-based action recognition.
\newblock In \emph{Humanoids}.

\bibitem[{Cen et~al.(2023)Cen, Luan, Zhang, Pei, Zhang, Zhao, Shen, and Chen}]{cen2023devil}
Cen, J.; Luan, D.; Zhang, S.; Pei, Y.; Zhang, Y.; Zhao, D.; Shen, S.; and Chen, Q. 2023.
\newblock The devil is in the wrongly-classified samples: Towards unified open-set recognition.
\newblock \emph{arXiv preprint arXiv:2302.04002}.

\bibitem[{Chen et~al.(2022)Chen, Peng, Wang, and Tian}]{chen2021adversarial}
Chen, G.; Peng, P.; Wang, X.; and Tian, Y. 2022.
\newblock Adversarial reciprocal points learning for open set recognition.
\newblock \emph{TPAMI}.

\bibitem[{Chen et~al.(2020)Chen, Qiao, Shi, Peng, Li, Huang, Pu, and Tian}]{chen2020learning}
Chen, G.; Qiao, L.; Shi, Y.; Peng, P.; Li, J.; Huang, T.; Pu, S.; and Tian, Y. 2020.
\newblock Learning open set network with discriminative reciprocal points.
\newblock In \emph{ECCV}.

\bibitem[{Chen et~al.(2021)Chen, Zhang, Yuan, Li, Deng, and Hu}]{chen2021channel}
Chen, Y.; Zhang, Z.; Yuan, C.; Li, B.; Deng, Y.; and Hu, W. 2021.
\newblock Channel-wise topology refinement graph convolution for skeleton-based action recognition.
\newblock In \emph{ICCV}.

\bibitem[{Cheng et~al.(2020)Cheng, Zhang, Cao, Shi, Cheng, and Lu}]{cheng2020decoupling}
Cheng, K.; Zhang, Y.; Cao, C.; Shi, L.; Cheng, J.; and Lu, H. 2020.
\newblock Decoupling gcn with dropgraph module for skeleton-based action recognition.
\newblock In \emph{ECCV}.

\bibitem[{Dai et~al.(2023)Dai, Das, Sharma, Minciullo, Garattoni, Bremond, and Francesca}]{dai2022toyota}
Dai, R.; Das, S.; Sharma, S.; Minciullo, L.; Garattoni, L.; Bremond, F.; and Francesca, G. 2023.
\newblock Toyota smarthome untrimmed: Real-world untrimmed videos for activity detection.
\newblock \emph{TPAMI}.

\bibitem[{Ding et~al.(2023)Ding, Liang, Perozzi, Chen, Wang, Hong, Chi, Liu, and Cheng}]{ding2023hyperformer}
Ding, K.; Liang, A.~J.; Perozzi, B.; Chen, T.; Wang, R.; Hong, L.; Chi, E.~H.; Liu, H.; and Cheng, D.~Z. 2023.
\newblock {HyperFormer:} {Learning} expressive sparse feature representations via hypergraph transformer.
\newblock In \emph{SIGIR}.

\bibitem[{Du et~al.(2023)Du, Shringi, Hoogs, and Funk}]{du2023reconstructing}
Du, D.; Shringi, A.; Hoogs, A.; and Funk, C. 2023.
\newblock Reconstructing humpty dumpty: Multi-feature graph autoencoder for open set action recognition.
\newblock In \emph{WACV}.

\bibitem[{Duan et~al.(2022)Duan, Zhao, Chen, Lin, and Dai}]{duan2022revisiting}
Duan, H.; Zhao, Y.; Chen, K.; Lin, D.; and Dai, B. 2022.
\newblock Revisiting skeleton-based action recognition.
\newblock In \emph{CVPR}.

\bibitem[{Fontanel et~al.(2020)Fontanel, Cermelli, Mancini, Bulo, Ricci, and Caputo}]{fontanel2020boosting}
Fontanel, D.; Cermelli, F.; Mancini, M.; Bulo, S.~R.; Ricci, E.; and Caputo, B. 2020.
\newblock Boosting deep open world recognition by clustering.
\newblock \emph{RA-L}.

\bibitem[{Geng and Chen(2020)}]{geng2020collective}
Geng, C.; and Chen, S. 2020.
\newblock Collective decision for open set recognition.
\newblock \emph{TKDE}.

\bibitem[{Gretton et~al.(2012)Gretton, Borgwardt, Rasch, Sch{\"o}lkopf, and Smola}]{gretton2012kernel}
Gretton, A.; Borgwardt, K.~M.; Rasch, M.~J.; Sch{\"o}lkopf, B.; and Smola, A. 2012.
\newblock A kernel two-sample test.
\newblock \emph{JMLR}.

\bibitem[{Hendrycks and Gimpel(2017)}]{hendrycks2016baseline}
Hendrycks, D.; and Gimpel, K. 2017.
\newblock A baseline for detecting misclassified and out-of-distribution examples in neural networks.
\newblock \emph{ICLR}.

\bibitem[{Ke et~al.(2017)Ke, Bennamoun, An, Sohel, and Boussaid}]{ke2017new}
Ke, Q.; Bennamoun, M.; An, S.; Sohel, F.; and Boussaid, F. 2017.
\newblock A new representation of skeleton sequences for 3D action recognition.
\newblock In \emph{ICCV}.

\bibitem[{Kipf and Welling(2016)}]{kipf2016semi}
Kipf, T.~N.; and Welling, M. 2016.
\newblock Semi-supervised classification with graph convolutional networks.
\newblock \emph{arXiv preprint arXiv:1609.02907}.

\bibitem[{Krishnan, Subedar, and Tickoo(2018)}]{krishnan2018bar}
Krishnan, R.; Subedar, M.; and Tickoo, O. 2018.
\newblock BAR: Bayesian activity recognition using variational inference.
\newblock \emph{arXiv preprint arXiv:1811.03305}.

\bibitem[{Lee et~al.(2022)Lee, Lee, Lee, and Lee}]{lee2022hierarchically}
Lee, J.; Lee, M.; Lee, D.; and Lee, S. 2022.
\newblock Hierarchically decomposed graph convolutional networks for skeleton-based action recognition.
\newblock \emph{arXiv preprint arXiv:2208.10741}.

\bibitem[{Liang et~al.(2019)Liang, Yang, Deng, Wang, and Wang}]{liang2019hierarchical}
Liang, Z.; Yang, M.; Deng, L.; Wang, C.; and Wang, B. 2019.
\newblock Hierarchical depthwise graph convolutional neural network for {3D} semantic segmentation of point clouds.
\newblock In \emph{ICRA}.

\bibitem[{Liu et~al.(2020)Liu, Shahroudy, Perez, Wang, Duan, and Kot}]{liu2020ntu}
Liu, J.; Shahroudy, A.; Perez, M.; Wang, G.; Duan, L.; and Kot, A.~C. 2020.
\newblock {NTU RGB+D 120:} {A} large-scale benchmark for {3D} human activity understanding.
\newblock \emph{TPAMI}.

\bibitem[{Liu, Liu, and Chen(2017)}]{liu2017enhanced}
Liu, M.; Liu, H.; and Chen, C. 2017.
\newblock Enhanced skeleton visualization for view invariant human action recognition.
\newblock \emph{PR}.

\bibitem[{Lu et~al.(2022)Lu, Xu, Li, Cheng, and Niu}]{lu2022pmal}
Lu, J.; Xu, Y.; Li, H.; Cheng, Z.; and Niu, Y. 2022.
\newblock {PMAL:} {Open} set recognition via robust prototype mining.
\newblock In \emph{AAAI}.

\bibitem[{Meyer and Drummond(2019)}]{meyer2019importance}
Meyer, B.~J.; and Drummond, T. 2019.
\newblock The importance of metric learning for robotic vision: Open set recognition and active learning.
\newblock In \emph{ICRA}.

\bibitem[{Miller et~al.(2018)Miller, Nicholson, Dayoub, and S{\"u}nderhauf}]{miller2018dropout}
Miller, D.; Nicholson, L.; Dayoub, F.; and S{\"u}nderhauf, N. 2018.
\newblock Dropout sampling for robust object detection in open-set conditions.
\newblock In \emph{ICRA}.

\bibitem[{Oza and Patel(2019)}]{oza2019c2ae}
Oza, P.; and Patel, V.~M. 2019.
\newblock C2AE: Class conditioned auto-encoder for open-set recognition.
\newblock In \emph{CVPR}.

\bibitem[{Plizzari, Cannici, and Matteucci(2021)}]{plizzari2021spatial}
Plizzari, C.; Cannici, M.; and Matteucci, M. 2021.
\newblock Spatial temporal transformer network for skeleton-based action recognition.
\newblock In \emph{ICPRW}.

\bibitem[{Roitberg et~al.(2020)Roitberg, Ma, Haurilet, and Stiefelhagen}]{roitberg2020open}
Roitberg, A.; Ma, C.; Haurilet, M.; and Stiefelhagen, R. 2020.
\newblock Open set driver activity recognition.
\newblock In \emph{IV}.

\bibitem[{Scheirer et~al.(2013)Scheirer, de~Rezende~Rocha, Sapkota, and Boult}]{scheirer2012toward}
Scheirer, W.~J.; de~Rezende~Rocha, A.; Sapkota, A.; and Boult, T.~E. 2013.
\newblock Toward open set recognition.
\newblock \emph{TPAMI}.

\bibitem[{Shahroudy et~al.(2016)Shahroudy, Liu, Ng, and Wang}]{shahroudy2016ntu}
Shahroudy, A.; Liu, J.; Ng, T.-T.; and Wang, G. 2016.
\newblock {NTU RGB+D:} {A} large scale dataset for {3D} human activity analysis.
\newblock In \emph{CVPR}.

\bibitem[{Shi et~al.(2019)Shi, Zhang, Cheng, and Lu}]{shi2019two}
Shi, L.; Zhang, Y.; Cheng, J.; and Lu, H. 2019.
\newblock Two-stream adaptive graph convolutional networks for skeleton-based action recognition.
\newblock In \emph{CVPR}.

\bibitem[{Shi et~al.(2020)Shi, Zhang, Cheng, and Lu}]{shi2020decoupled}
Shi, L.; Zhang, Y.; Cheng, J.; and Lu, H. 2020.
\newblock Decoupled spatial-temporal attention network for skeleton-based action-gesture recognition.
\newblock In \emph{ACCV}.

\bibitem[{Shi(2023)}]{shi2023open}
Shi, Y. 2023.
\newblock Open set action recognition based on skeleton.
\newblock In \emph{ICCCS}.

\bibitem[{Shu et~al.(2018)Shu, Shi, Wang, Zou, Yuan, and Tian}]{shu2018odn}
Shu, Y.; Shi, Y.; Wang, Y.; Zou, Y.; Yuan, Q.; and Tian, Y. 2018.
\newblock {ODN:} {Opening} the deep network for open-set action recognition.
\newblock In \emph{ICME}.

\bibitem[{Subedar et~al.(2019)Subedar, Krishnan, Meyer, Tickoo, and Huang}]{subedar2019uncertainty}
Subedar, M.; Krishnan, R.; Meyer, P.~L.; Tickoo, O.; and Huang, J. 2019.
\newblock Uncertainty-aware audiovisual activity recognition using deep bayesian variational inference.
\newblock In \emph{ICCV}.

\bibitem[{Sun et~al.(2020)Sun, Yang, Zhang, Ling, and Peng}]{sun2020conditional}
Sun, X.; Yang, Z.; Zhang, C.; Ling, K.-V.; and Peng, G. 2020.
\newblock Conditional Gaussian distribution learning for open set recognition.
\newblock In \emph{CVPR}.

\bibitem[{van~der Maaten and Hinton(2008)}]{JMLR:v9:vandermaaten08a}
van~der Maaten, L.; and Hinton, G. 2008.
\newblock Visualizing Data using t-SNE.
\newblock \emph{JMLR}.

\bibitem[{Xin et~al.(2023)Xin, Liu, Liu, Chen, Yu, and Miao}]{xin2023transformer}
Xin, W.; Liu, R.; Liu, Y.; Chen, Y.; Yu, W.; and Miao, Q. 2023.
\newblock Transformer for skeleton-based action recognition: A review of recent advances.
\newblock \emph{Neurocomputing}.

\bibitem[{Yan, Xiong, and Lin(2018)}]{yan2018spatial}
Yan, S.; Xiong, Y.; and Lin, D. 2018.
\newblock Spatial temporal graph convolutional networks for skeleton-based action recognition.
\newblock In \emph{AAAI}.

\bibitem[{Yang et~al.(2019)Yang, Hou, Lang, Guan, Huang, and Xu}]{yang2019open}
Yang, Y.; Hou, C.; Lang, Y.; Guan, D.; Huang, D.; and Xu, J. 2019.
\newblock Open-set human activity recognition based on micro-Doppler signatures.
\newblock \emph{PR}.

\bibitem[{Ye et~al.(2020)Ye, Pu, Zhong, Li, Xie, and Tang}]{ye2020dynamic}
Ye, F.; Pu, S.; Zhong, Q.; Li, C.; Xie, D.; and Tang, H. 2020.
\newblock Dynamic GCN: Context-enriched topology learning for skeleton-based action recognition.
\newblock In \emph{MM}.

\bibitem[{Yoshihashi et~al.(2019)Yoshihashi, Shao, Kawakami, You, Iida, and Naemura}]{yoshihashi2019classification}
Yoshihashi, R.; Shao, W.; Kawakami, R.; You, S.; Iida, M.; and Naemura, T. 2019.
\newblock Classification-reconstruction learning for open-set recognition.
\newblock In \emph{CVPR}.

\bibitem[{Zhou et~al.(2022)Zhou, Li, Cheng, Geng, Xie, and Keuper}]{zhou2022hypergraph}
Zhou, Y.; Li, C.; Cheng, Z.-Q.; Geng, Y.; Xie, X.; and Keuper, M. 2022.
\newblock Hypergraph transformer for skeleton-based action recognition.
\newblock \emph{arXiv preprint arXiv:2211.09590}.

\end{thebibliography}

\appendix
\section{Discussion of Societal Impacts and Limitation}
\noindent\textbf{Societal Impacts.} In our work, we construct the first large-scale open-set skeleton-based action recognition benchmarks by using diverse backbones, datasets, and evaluation settings, which is named the OS-SAR benchmark.
Seven well-established open-set recognition approaches, \ie, vanilla SoftMax~\cite{hendrycks2016baseline}, RPL~\cite{chen2020learning}, ARPL~\cite{chen2021adversarial}, DEAR~\cite{bao2021evidential}, Humpty Dumpty~\cite{du2023reconstructing}, PMAL~\cite{lu2022pmal}, and MCD-V~\cite{roitberg2020open} are leveraged to serve as baselines, which are derived from image classification task and the video-based human action recognition task due to the lack of OS-SAR research. Through our experiments, we found that the existing approaches can not show generally great performance on the OS-SAR benchmark due to the large discrepancy between the image/video data and sparse skeleton data.

To delve deeper into the underlying issue considering all the existing approaches, we find out that the in- and out-of-distribution samples can not disentangle well in terms of the predicted open-set probabilities. Keeping this problem in mind, we proposed CrossMax by using cross-modality mean max discrepancy during the training and cross-modality logits refinement technique during the testing together, which shows superior better disentanglement in terms of the open-set probability compared between in- and out-of-distribution samples. State-of-the-art OSSAR performances are delivered by our approach across backbones, datasets, and evaluation settings.
However, our method still has the potential to give misclassification, and biased content which may cause false predictions resulting in a negative social impact.
\begin{figure*}[htb!]
    \centering
    \begin{minipage}[t]{0.5\columnwidth}
        \includegraphics[width=0.99\linewidth]{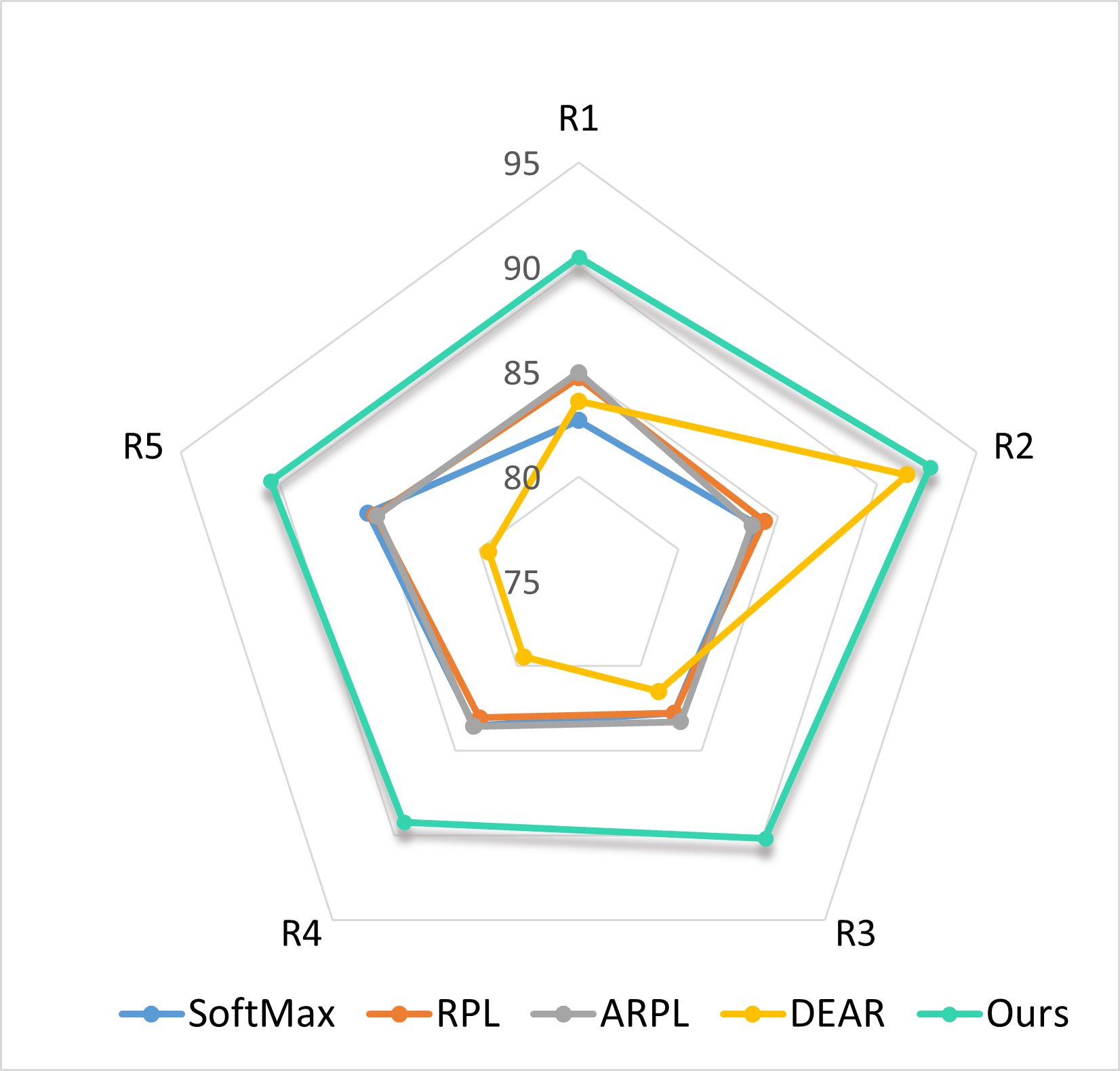}
        \centering
        \subcaption{O-AUROC for Case1}\label{score_a1}
    \end{minipage}%
    \begin{minipage}[t]{0.5\columnwidth}
    \includegraphics[width=0.99\linewidth]{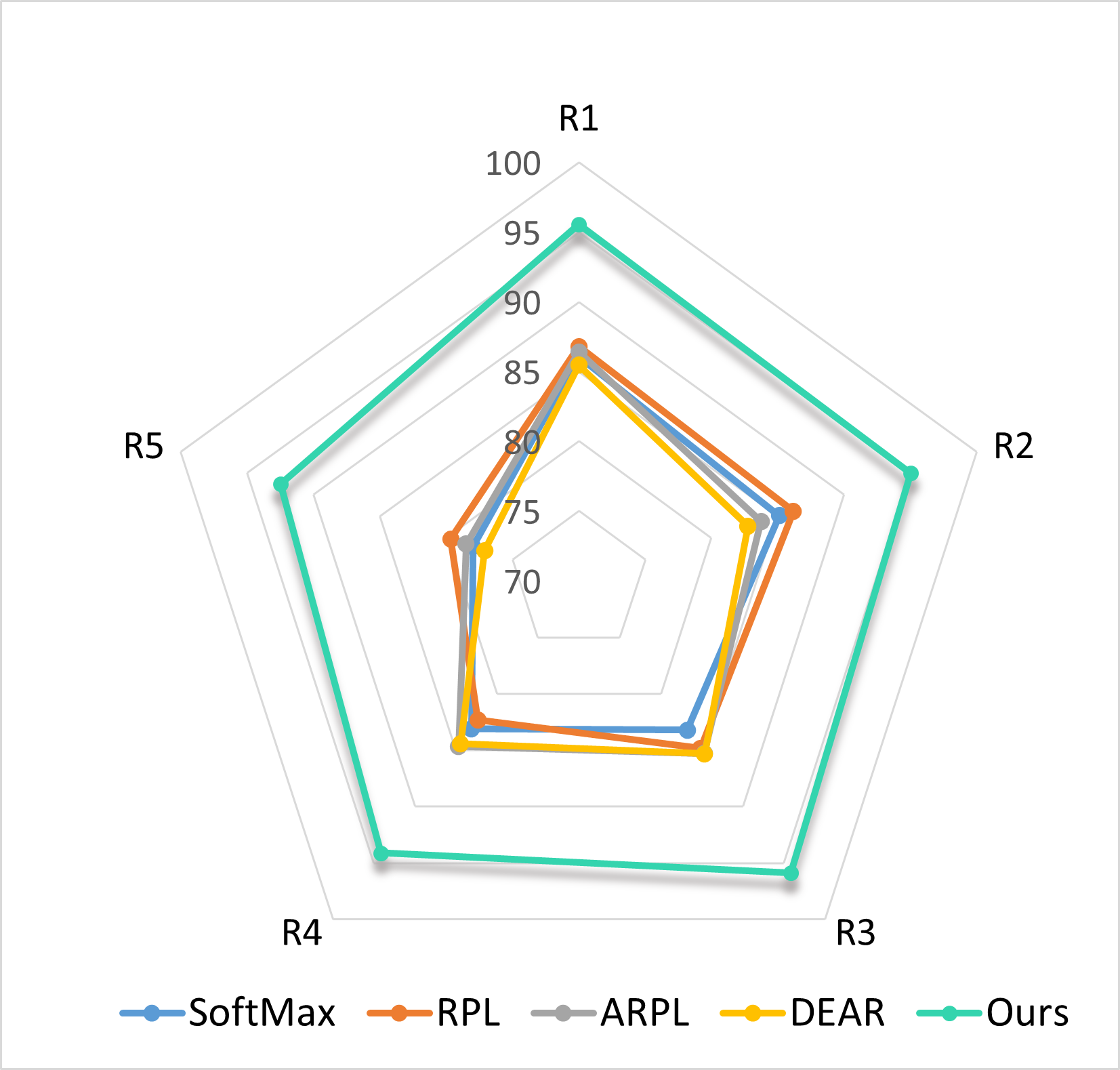}
        \centering
        \subcaption{O-AUROC for Case2}\label{score_b1}
    \end{minipage}%
        \begin{minipage}[t]{0.5\columnwidth}
    \includegraphics[width=0.99\linewidth]{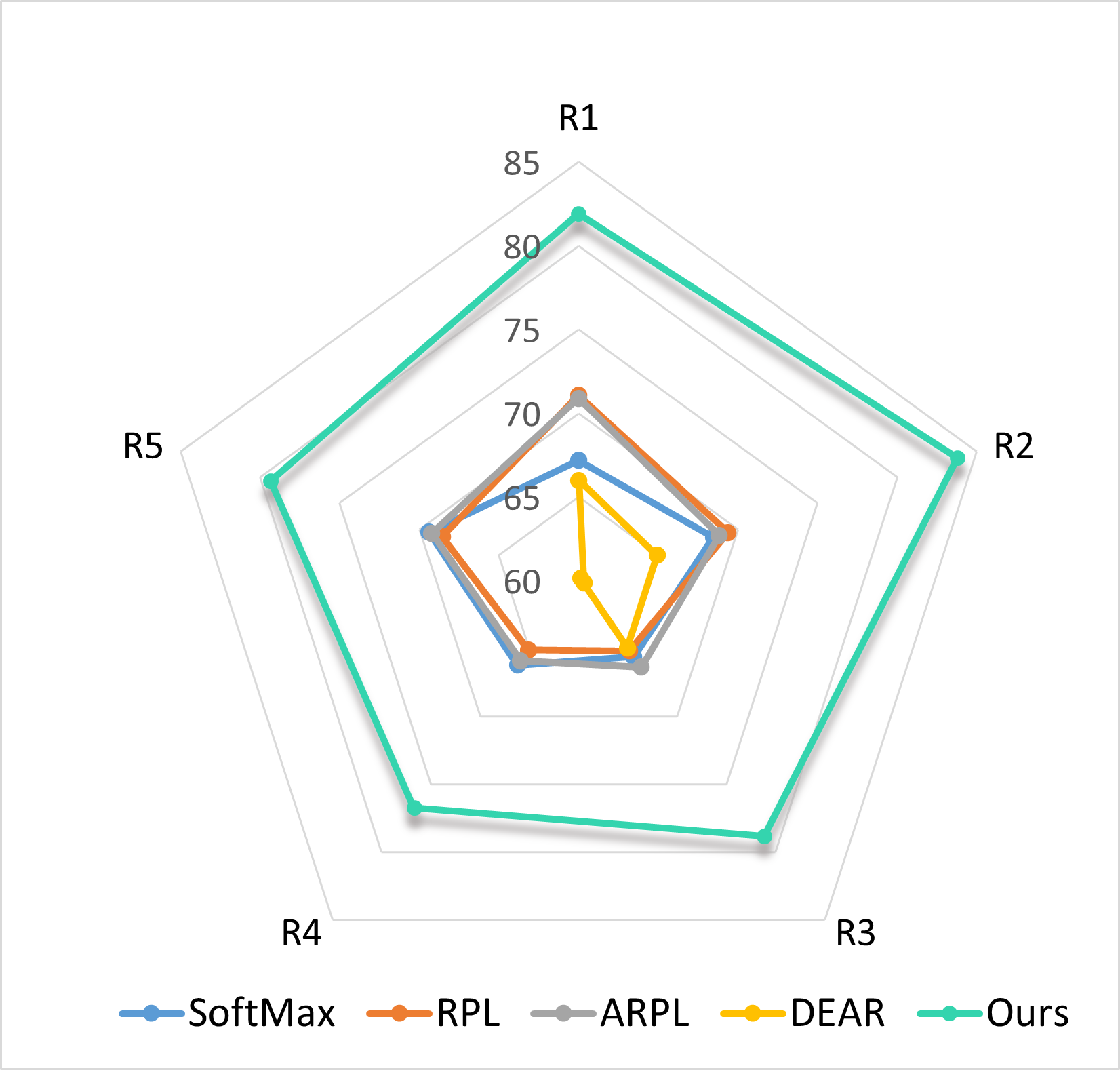}
        \centering
        \subcaption{O-AUPR for Case1}\label{score_b2}
    \end{minipage}%
        \begin{minipage}[t]{0.5\columnwidth}
    \includegraphics[width=0.99\linewidth]{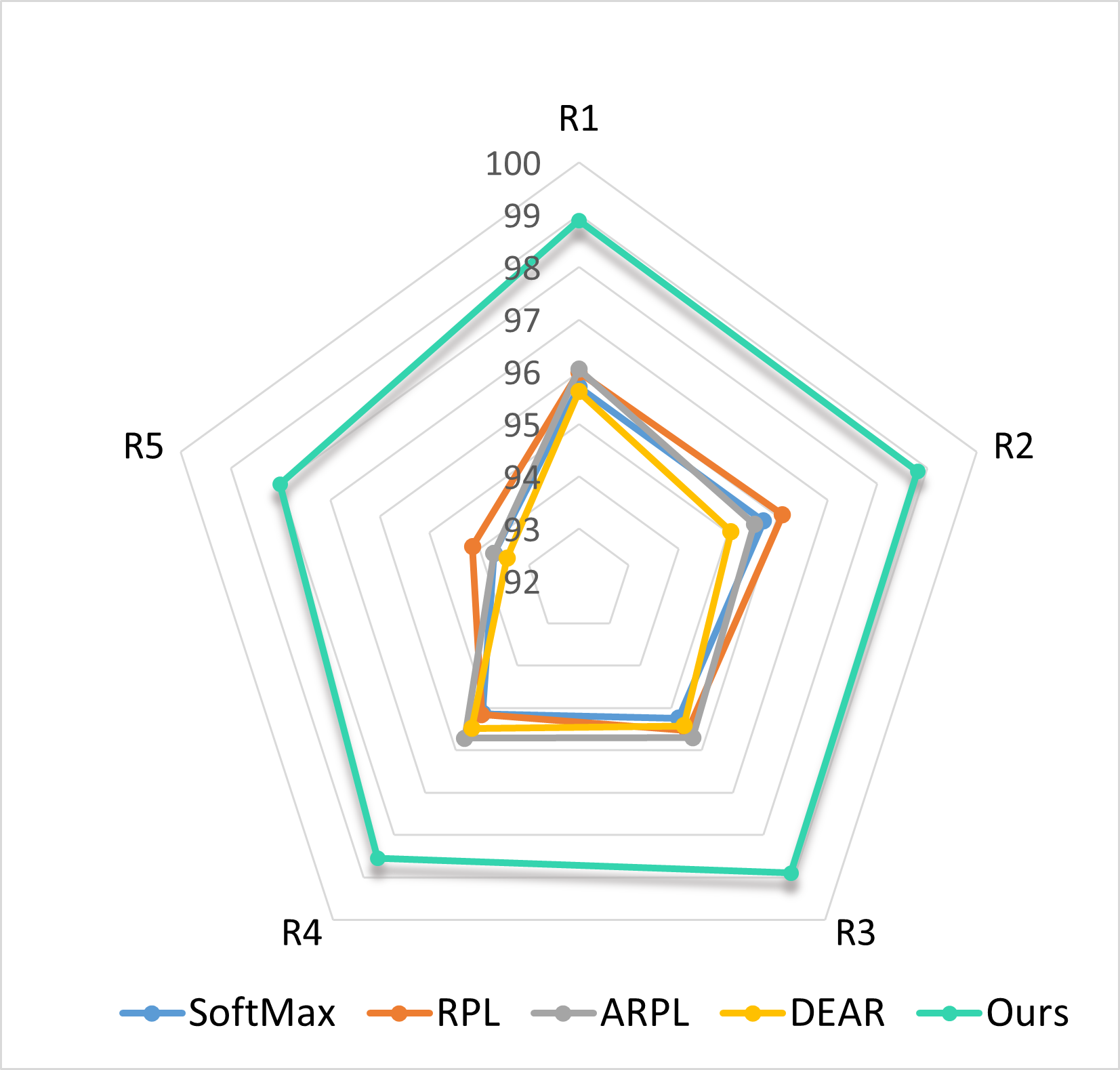}
        \centering
        \subcaption{O-AUPR for Case2}\label{score_b3}
    \end{minipage}%

    \caption{Comparison of open-set recognition performances using CTRGCN~\cite{chen2021channel} backbone on NTU60~\cite{shahroudy2016ntu} cross-subject evaluation for five different random splits for different open-set ratios, where R1 to R5 indicates the five random splits.}
    \label{fig:openset_score_dist_ablation_unseen}

\end{figure*}

\begin{figure*}[htb!]
    \centering
    \begin{minipage}[t]{0.5\columnwidth}
        \includegraphics[width=0.99\linewidth]{Figure/No-Noise-O-AUROC.png}
        \centering
        \subcaption{O-AUROC w/o Noise}\label{score_a4}
    \end{minipage}%
    \begin{minipage}[t]{0.5\columnwidth}
    \includegraphics[width=0.99\linewidth]{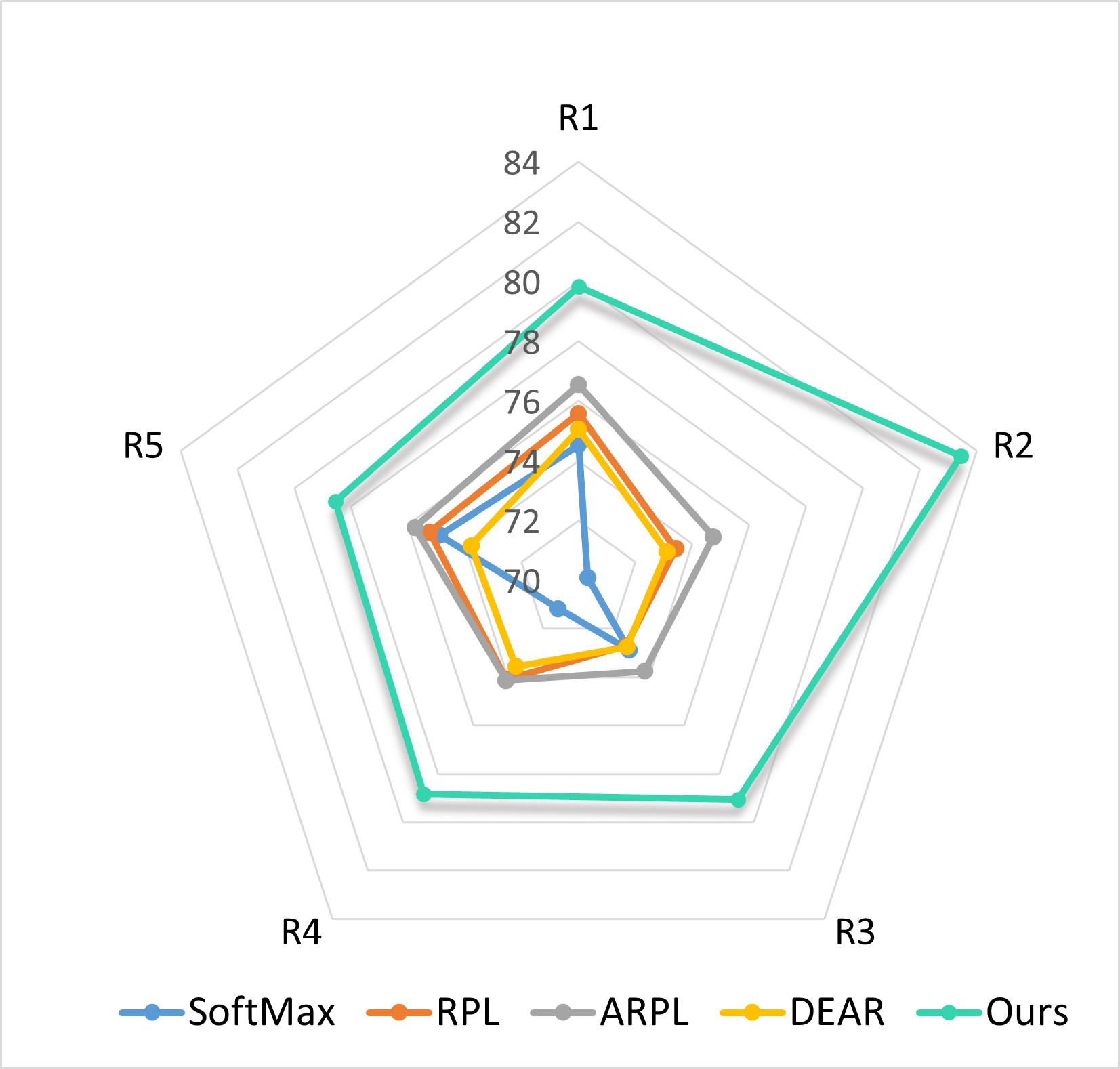}
        \centering
        \subcaption{O-AUROC w/ Noise}\label{score_b5}
    \end{minipage}%
        \begin{minipage}[t]{0.5\columnwidth}
    \includegraphics[width=0.99\linewidth]{Figure/No-Noise-O-AURPR.png}
        \centering
        \subcaption{O-AUPR w/o Noise}\label{score_b6}
    \end{minipage}%
        \begin{minipage}[t]{0.5\columnwidth}
    \includegraphics[width=0.99\linewidth]{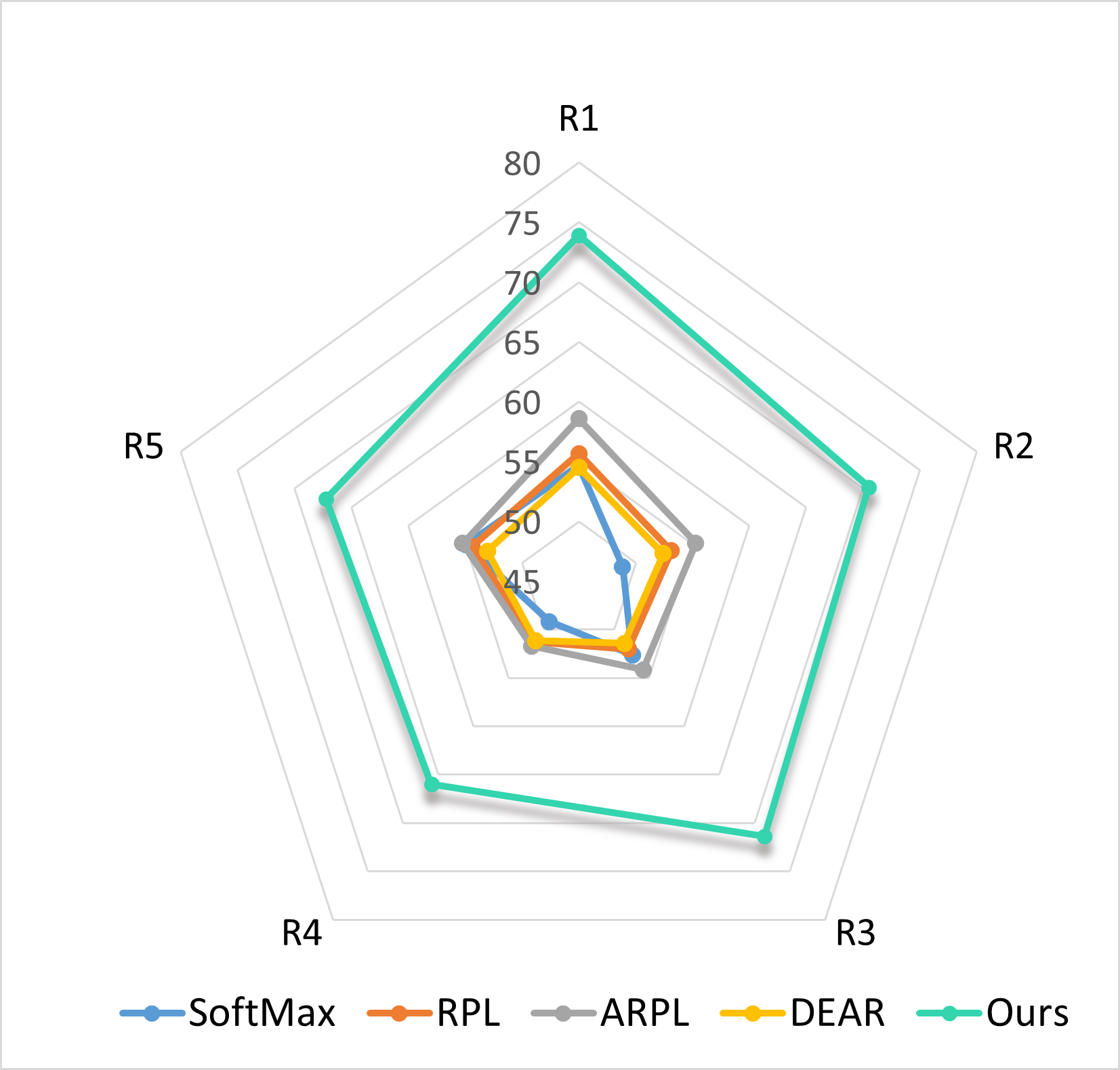}
        \centering
        \subcaption{O-AUPR w/ Noise}\label{score_b7}
    \end{minipage}%

    \caption{Comparison of open-set recognition performances using CTRGCN~\cite{chen2021channel} backbone on NTU60~\cite{shahroudy2016ntu} cross-subject evaluation for five different random splits for w/ noise and w/o noise scenarios, where R1 to R5 indicates the five random splits.}
    \label{fig:openset_score_dist_ablation_noise}

\end{figure*}
\begin{figure*}[htb!]
    \centering
    \begin{minipage}[t]{0.5\columnwidth}
        \includegraphics[width=0.99\linewidth]{Figure/No-Noise-O-AUROC.png}
        \centering
        \subcaption{O-AUROC w/o Occlusion}\label{score_a8}
    \end{minipage}%
    \begin{minipage}[t]{0.5\columnwidth}
    \includegraphics[width=0.99\linewidth]{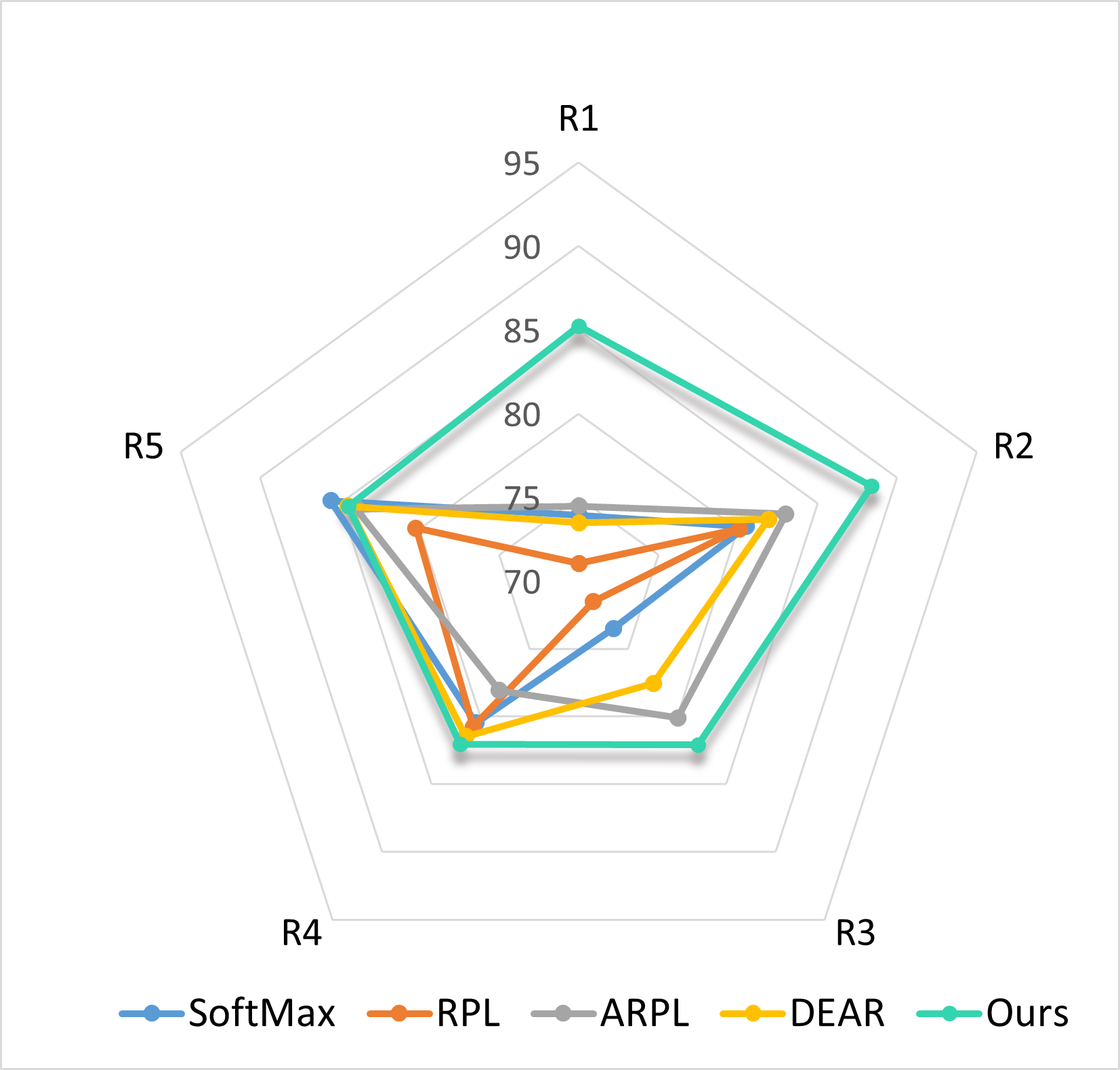}
        \centering
        \subcaption{O-AUROC w/ Occlusion}\label{score_b9}
    \end{minipage}%
        \begin{minipage}[t]{0.5\columnwidth}
    \includegraphics[width=0.99\linewidth]{Figure/No-Noise-O-AURPR.png}
        \centering
        \subcaption{O-AUPR w/o Occlusion}\label{score_b10}
    \end{minipage}%
        \begin{minipage}[t]{0.5\columnwidth}
    \includegraphics[width=0.99\linewidth]{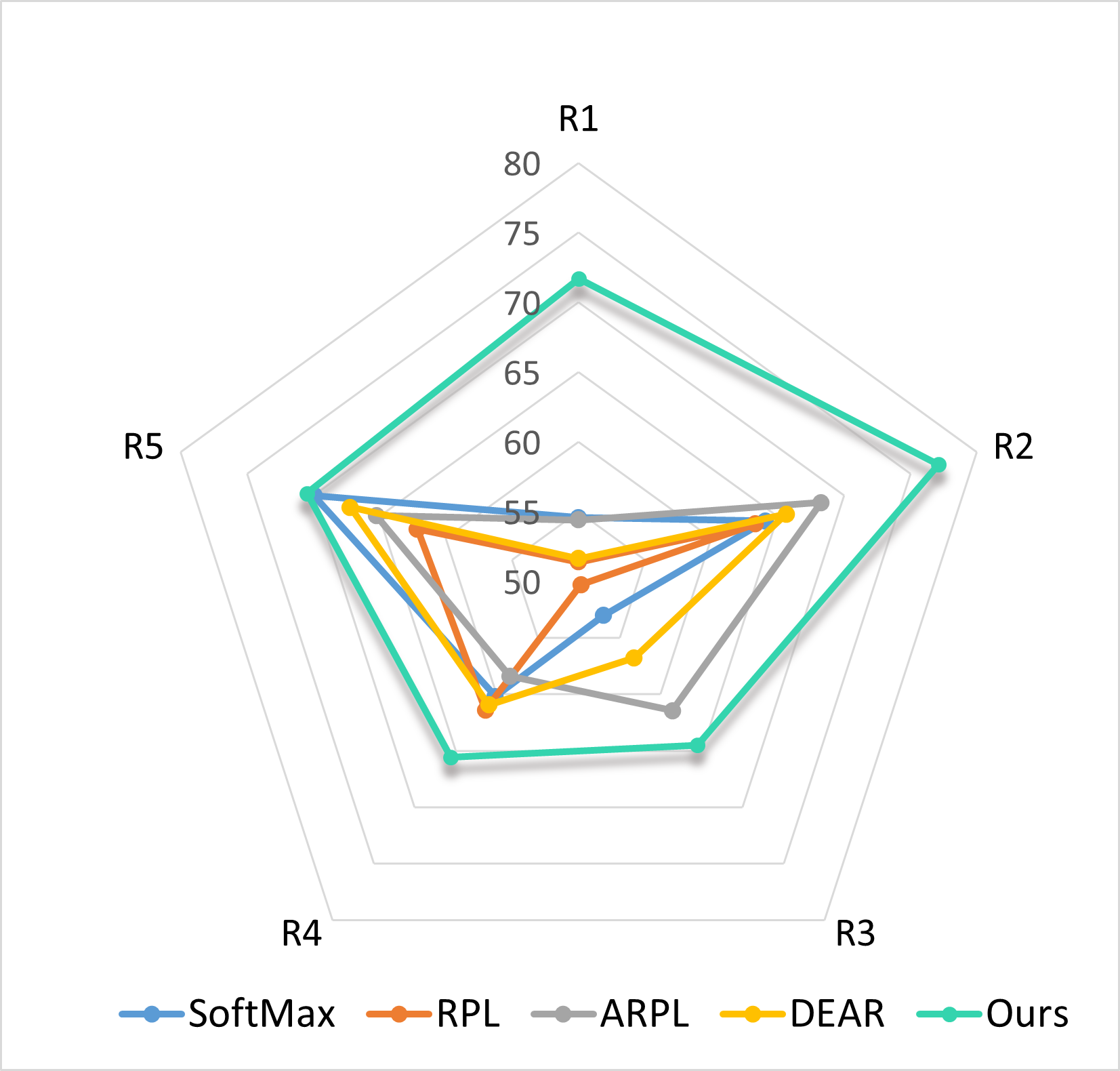}
        \centering
        \subcaption{O-AUPR w/ Occlusion}\label{score_b11}
    \end{minipage}%

    \caption{Comparison of open-set recognition performances using CTRGCN~\cite{chen2021channel} backbone on NTU60~\cite{shahroudy2016ntu} cross-subject evaluation for five different random splits for w/ occlusion and w/o occlusion scenarios, where R1 to R5 indicates the five random splits.}
    \label{fig:openset_score_dist_ablation_occ}

\end{figure*}
\noindent\textbf{Limitations.} Our method relies on ensemble modalities which need to triplet the model three times. However, due to the small size of the GCN models designed uniquely for skeleton data, the usage of memory is still acceptable compared with the models leveraged in image/video-based tasks.
\begin{figure*}[htb!]
\centering
\includegraphics[width=1\linewidth]{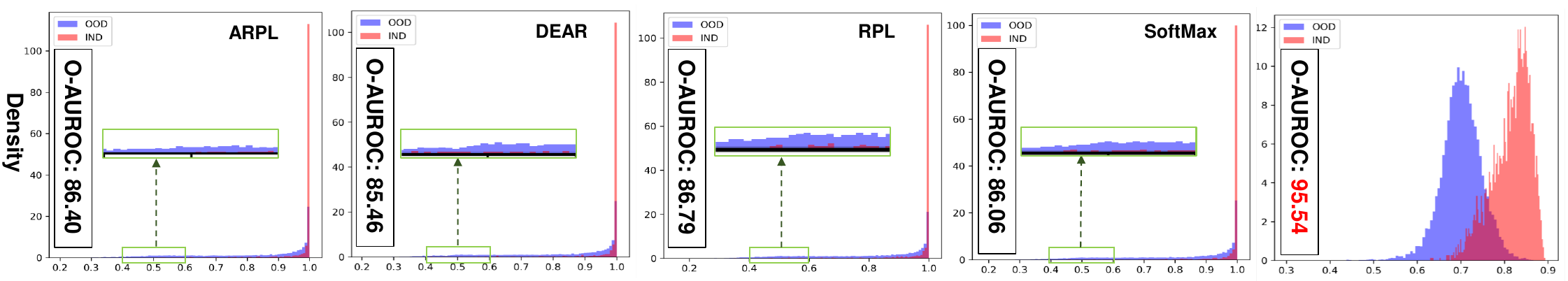}

\caption{Comparison of the open-set probabilities using CTRGCN~\cite{chen2021channel} as the backbone on NTU60~\cite{shahroudy2016ntu} for cross-subject evaluation for one random split (Run1) for Case2 open-set ratio. Our approach achieves better disentanglement of the in- and out-of-distribution samples. The results of CrossMax are shown on the far right.}

\label{fig:probability_score_unseen10}
\end{figure*}

\begin{figure*}[htb!]
\centering
\includegraphics[width=1\linewidth]{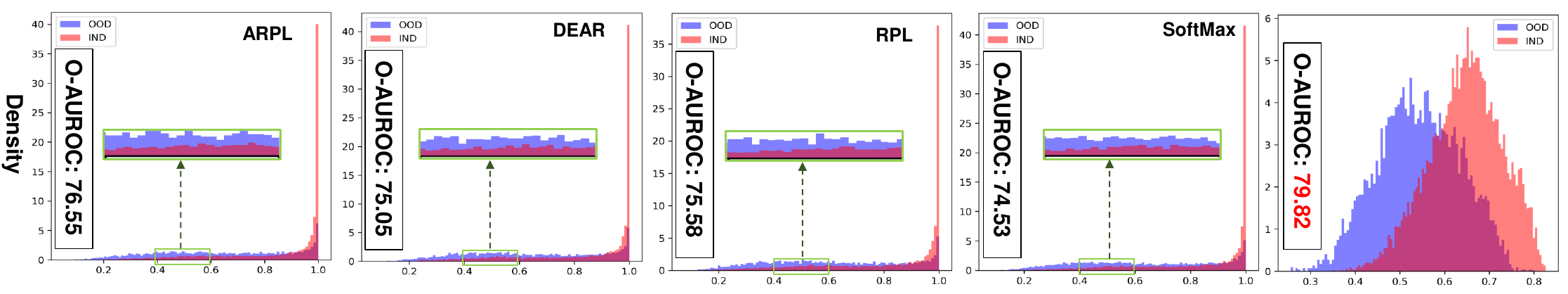}

\caption{Comparison of the open-set probabilities using CTRGCN~\cite{chen2021channel} as the backbone on NTU60~\cite{shahroudy2016ntu} for cross-subject evaluation for one random split (Run1) under Gaussian noise disturbance. Our approach achieves better disentanglement of the in- and out-of-distribution samples. The results of CrossMax are shown on the far right.}

\label{fig:probability_score_noise}
\end{figure*}

\begin{figure*}[htb!]
\centering
\includegraphics[width=1\linewidth]{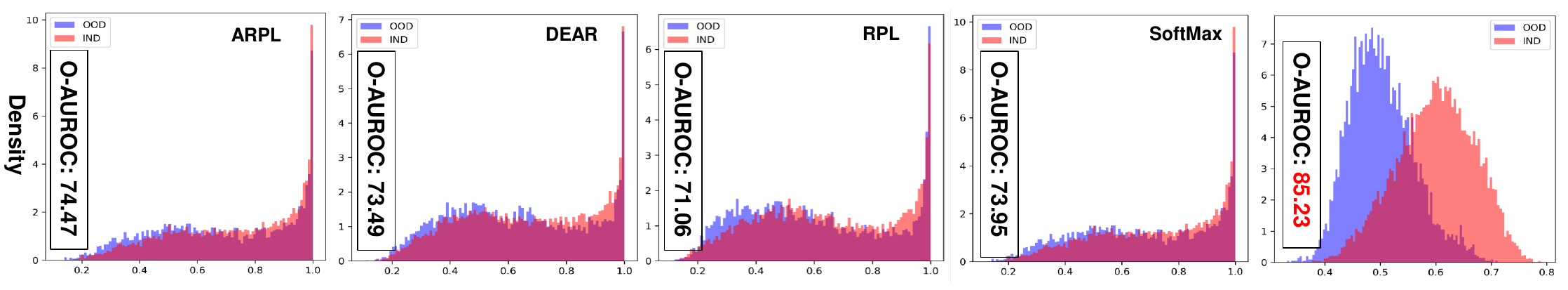}

\caption{Comparison of the open-set probabilities using CTRGCN~\cite{chen2021channel} as the backbone on NTU60~\cite{shahroudy2016ntu} for cross-subject evaluation for one random split (Run1) under random occlusion disturbance. Our approach achieves better disentanglement of the in- and out-of-distribution samples. The results of CrossMax are shown on the far right.}

\label{fig:probability_score_occ}
\end{figure*}
\begin{figure*}[p!]
	\centering
		\begin{minipage}[t]{1\columnwidth}
			\includegraphics[width=0.8\linewidth]{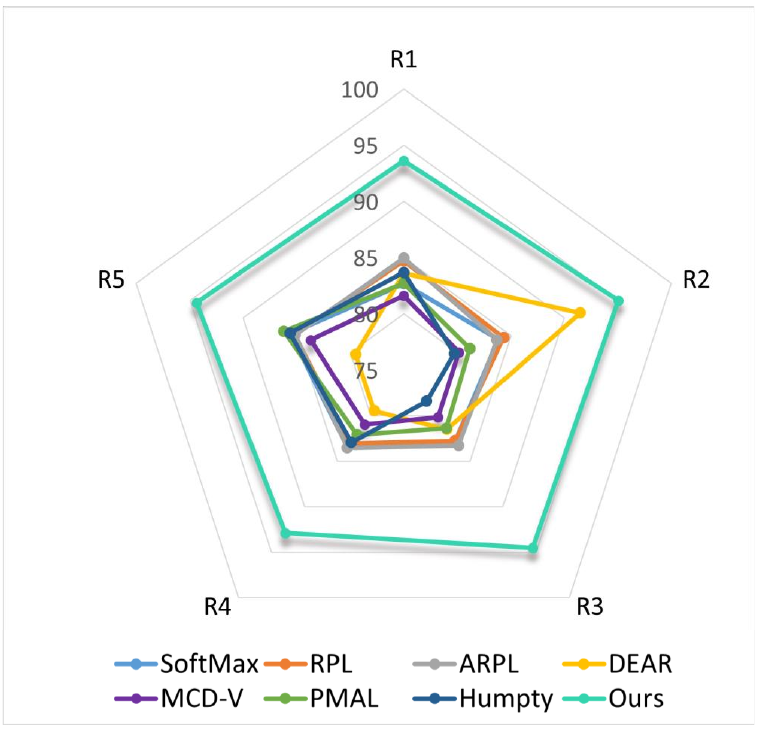} 
            \label{fig:all_run_1}
            \subcaption{O-AUROC (CTRGCN)}
		\end{minipage}
		\label{fig:grid_4figs_1cap_4subcap_1}
    		\begin{minipage}[t]{1\columnwidth}
   		 	\includegraphics[width=0.8\linewidth]{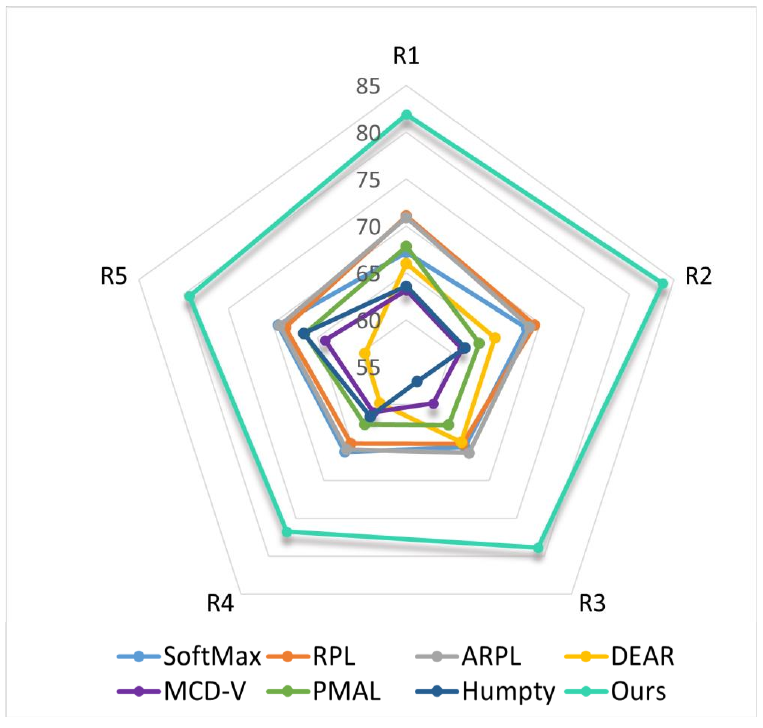}
            \label{fig:all_run_2}
            \subcaption{O-AUPR (CTRGCN)}
    		\end{minipage}
		\label{fig:grid_4figs_1cap_4subcap_2}\\
		\begin{minipage}[t]{1\columnwidth}
			\includegraphics[width=0.8\linewidth]{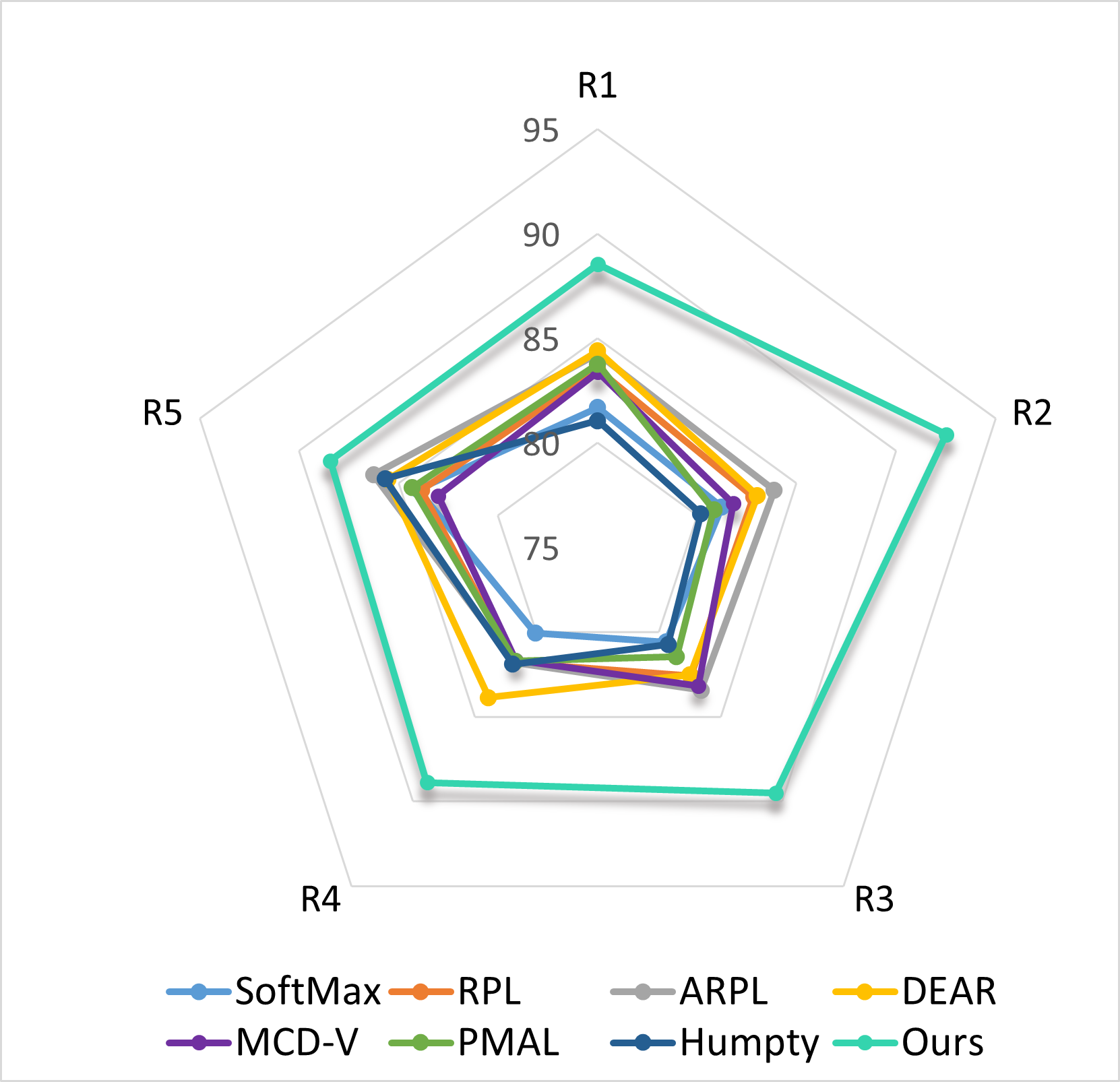}
            \label{fig:all_run_3}
            \subcaption{O-AUROC (HDGCN)}
		\end{minipage}
		\label{fig:grid_4figs_1cap_4subcap_3}
    		\begin{minipage}[t]{1\columnwidth}
		 	\includegraphics[width=0.8\linewidth]{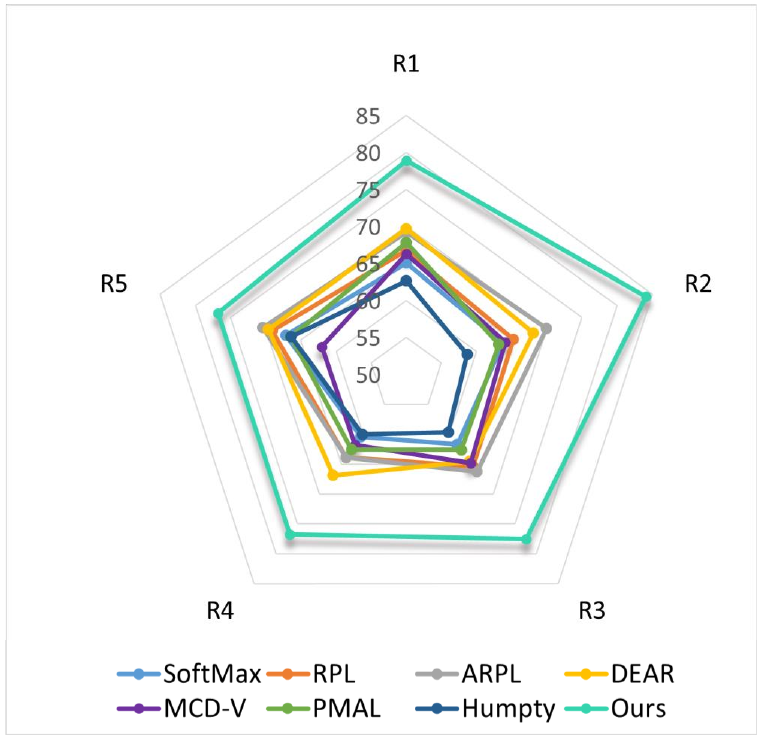}
            \label{fig:all_run_4}
            \subcaption{O-AUPR (HDGCN)}
    		\end{minipage}
		\label{fig:grid_4figs_1cap_4subcap_4}\\
		\begin{minipage}[t]{1\columnwidth}
			\includegraphics[width=0.8\linewidth]{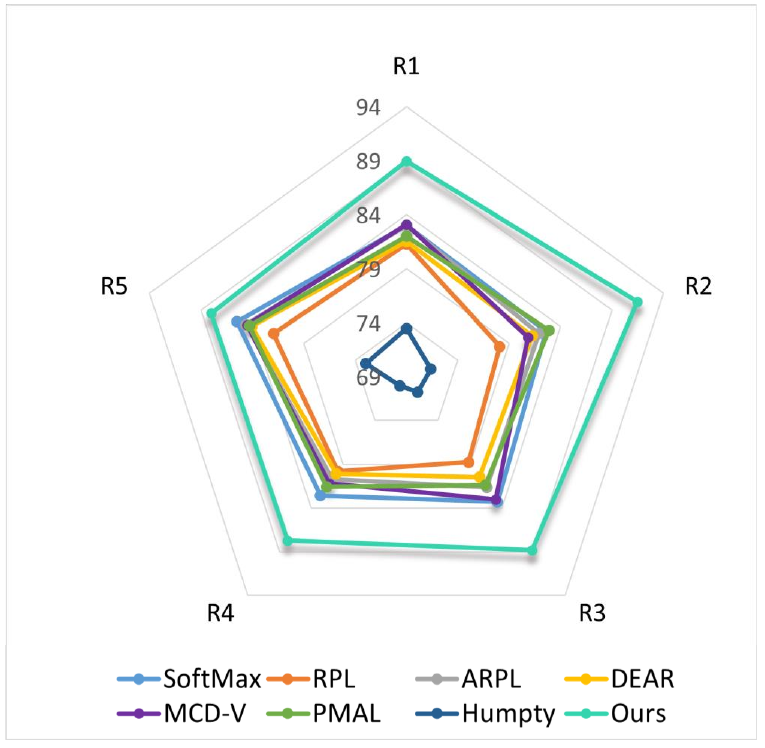} 
               \label{fig:all_run_5}
            \subcaption{O-AUROC (Hyperformer)}
		\end{minipage}
		\label{fig:grid_4figs_1cap_4subcap_5}
    		\begin{minipage}[t]{1\columnwidth}
		 	\includegraphics[width=0.8\linewidth]{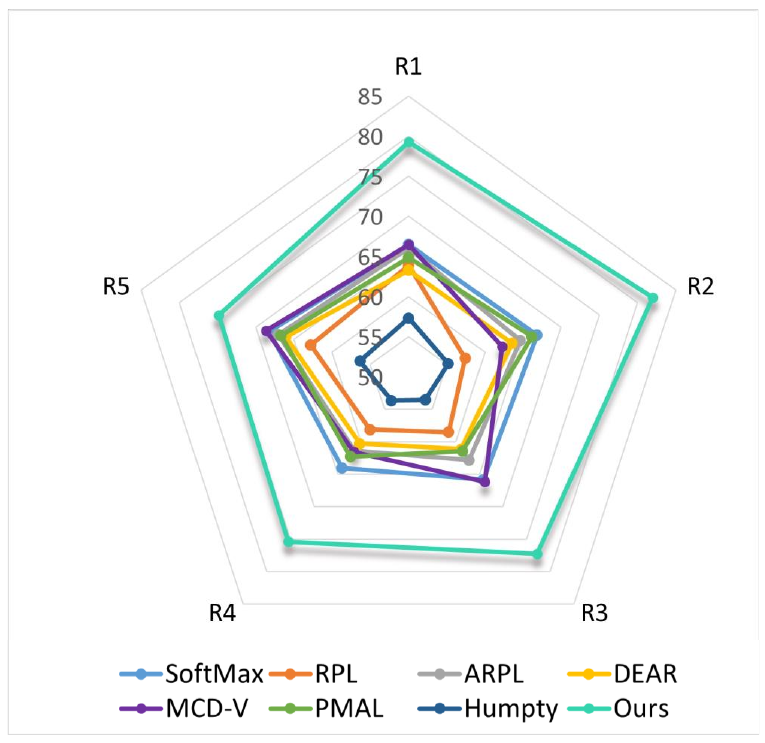}
            \label{fig:all_run_6}
            \subcaption{O-AUPR (Hyperformer)}
    		\end{minipage}
		\label{fig:grid_4figs_1cap_4subcap_6}
	\caption{Experimental results for all five random splits on NTU60~\cite{shahroudy2016ntu} dataset under cross-subject evaluation while considering the generalizability across feature extraction backbones, \ie, CTRGCN~\cite{chen2021channel}, HDGCN~\cite{liang2019hierarchical}, and Hyperformer~\cite{ding2023hyperformer}.}
	\label{fig:backbone}
\end{figure*}
\noindent\textbf{Future works.} Our empirical analysis reveals that the incorporation of multi-modality data yields enhanced open-set performance within the OS-SAR benchmark. As a consequence, we identify a compelling avenue for future research stemming from our devised OS-SAR benchmark. This prospective direction pertains to the optimal utilization of multi-modality data to amplify the efficacy of OS-SAR.
\begin{table}[t!]
\centering
\caption{Ablation study for OS-SAR under different open-set ratios using CTRGCN on NTU60~\cite{shahroudy2016ntu} dataset for cross-view and cross-subject evaluations, where the results are averaged on five random splits.}
\label{tab:openess}
\scalebox{0.90}{\begin{tabular}{lllllll} 
\toprule\midrule
\multirow{2}{*}{\textbf{Method}} & \multicolumn{2}{l}{\textbf{O-AUROC}} & \multicolumn{2}{l}{\textbf{O\_AUPR}} & \multicolumn{2}{l}{\textbf{C-ACC}} \\ 
\cline{2-7}
 & \textbf{CS} & \textbf{CV} & \textbf{CS} & \textbf{CV} & \textbf{CS} & \textbf{CV} \\ 
 \midrule
 \multicolumn{5}{l}{\textbf{\circled{1} Case1.}} \\
\midrule
SoftMax & 83.68 & 87.77 & 67.37 & 76.38 & 90.56 & 93.83 \\
RPL& 84.02 & 88.06 & 67.86 & 76.75 & 90.82 & 95.38  \\
ARPL  & 84.13 & 88.37 & 68.24 & 76.58 & 91.00 & 95.45 \\
DEAR & 83.11 & 87.54 & 63.07 & 75.52 & 84.14 & 95.41 \\ 
\midrule
{\cellcolor{LightCyan}}Ours &{\cellcolor{LightCyan}}\textbf{90.62} &{\cellcolor{LightCyan}}\textbf{94.14} &{\cellcolor{LightCyan}}\textbf{80.32} &{\cellcolor{LightCyan}}\textbf{88.07} & {\cellcolor{LightCyan}}\textbf{93.68} & {\cellcolor{LightCyan}}\textbf{97.51}  \\
\midrule
\multicolumn{5}{l}{\textbf{\circled{2} Case2.}} \\
\midrule
SoftMax & 83.10 & 85.58 & 91.54 & 96.05 & \textbf{95.10} & 95.76 \\
RPL &83.72 & 87.58 & 95.34& 96.51& 92.20 & 95.54 \\
ARPL & 83.72 & 87.52 & 95.34 & 96.49 & 92.20 & 95.49 \\
DEAR & 83.00 & 86.13 & 95.00 & 96.15 & 91.30 &  95.28\\ 
\midrule
{\cellcolor{LightCyan}}Ours&{\cellcolor{LightCyan}}\textbf{94.61} & {\cellcolor{LightCyan}}\textbf{96.20} & {\cellcolor{LightCyan}}\textbf{98.63} & {\cellcolor{LightCyan}}\textbf{99.04} & {\cellcolor{LightCyan}}94.17 & {\cellcolor{LightCyan}}\textbf{96.90} \\
\midrule\bottomrule
\end{tabular}}
\end{table}
\begin{table}[htb!]
\centering
\caption{Ablation study for OS-SAR under Gaussian noise disturbance using CTRGCN on NTU60~\cite{shahroudy2016ntu} dataset for cross-view and cross-subject evaluations, where the results are averaged on five random splits.}
\label{tab:noise}
\scalebox{0.90}{\begin{tabular}{lllllll} 
\toprule\midrule
\multirow{2}{*}{\textbf{Method}} & \multicolumn{2}{l}{\textbf{O-AUROC}} & \multicolumn{2}{l}{\textbf{O\_AUPR}} & \multicolumn{2}{l}{\textbf{C-ACC}} \\ 
\cline{2-7}
 & \textbf{CS} & \textbf{CV} & \textbf{CS} & \textbf{CV} & \textbf{CS} & \textbf{CV} \\ 
 \midrule
 \multicolumn{5}{l}{\textbf{\circled{1} Without Noise.}} \\
\midrule
SoftMax & 83.68 & 87.77 & 67.37 & 76.38 & 90.56 & 93.83 \\
RPL& 84.02 & 88.06 & 67.86 & 76.75 & 90.82 & 95.38 \\
ARPL  & 84.13 & 88.37 & 68.24 & 76.58 & 91.00 & 95.45 \\
DEAR & 83.11 & 87.54 & 63.07 & 75.52 & 84.14 & 95.41 \\ 
\midrule
{\cellcolor{LightCyan}}  Ours &{\cellcolor{LightCyan}} \textbf{90.62} &{\cellcolor{LightCyan}} \textbf{94.14} &{\cellcolor{LightCyan}} \textbf{80.32} &{\cellcolor{LightCyan}} \textbf{88.07} & {\cellcolor{LightCyan}}\textbf{93.68} & {\cellcolor{LightCyan}}\textbf{97.51}  \\
\midrule
\multicolumn{5}{l}{\textbf{\circled{2} Gaussian Noise Disturbance.}} \\
\midrule
SoftMax & 72.76 & 76.36 & 52.04 & 57.63 & 56.44 & 57.68 \\
RPL & 74.21 & 77.79 & 53.31 & 59.49 & 71.74 & 75.36 \\
ARPL & 74.99 & 78.66 & 54.99 & 60.55 & 82.76 & 88.26 \\
DEAR & 73.65 & 76.83 & 52.51 & 57.32 & 82.14 & 86.63 \\ 
\midrule
{\cellcolor{LightCyan}} Ours & {\cellcolor{LightCyan}}\textbf{79.94} & {\cellcolor{LightCyan}}\textbf{83.36} & {\cellcolor{LightCyan}}\textbf{66.03} & {\cellcolor{LightCyan}}\textbf{73.91} & {\cellcolor{LightCyan}}\textbf{85.93}  & {\cellcolor{LightCyan}}\textbf{89.31}  \\
\midrule\bottomrule
\end{tabular}}
\end{table}

\section{Ablation for Different Open-Set Ratios}
We conduct ablation studies on the four baselines with outstanding OS-SAR performances, \ie, SoftMax~\cite{hendrycks2016baseline}, RPL~\cite{chen2020learning}, ARPL~\cite{chen2021adversarial}, DEAR~\cite{du2023reconstructing}, and our CrossMax on NTU60 by using CTR-GCN as the backbone, where the experimental results are reported in Tab.~\ref{tab:openess} for both cross-subject and cross-view evaluations considering different open-set ratios, where \textit{Case1} indicates that $40$ classes are leveraged during training and serve as seen classes while $20$ classes are used as unseen classes, \textit{Case2} indicates that $10$ classes are leveraged during training and serve as seen classes while $50$ classes are used as unseen classes, which is more challenging due to the lack of a priori knowledge for more action categories during the training phase. The performances of the baselines for \textit{Case2} show a slight decay for O-AUROC, while our CrossMax shows comparable performances between different open-set ratios and achieves $10.89\%$ and $8.68\%$ improvements compared with ARPL in terms of the O-AUROC metric on cross-subject and cross-view evaluations, respectively.
Compared with the performance of our approach for \textit{Case1}, our approach achieves $3.99\%$ and $2.06\%$ improvements of O-AUROC and $18.31\%$ and $10.97\%$ improvements of O-AUPR for both the cross-subject and cross-view evaluations, demonstrating the fantastic performance of our approach when dealing with the challenging open-set scenario. We further deliver the predicted open-set probabilities for in- and out-of-distribution samples in Fig.~\ref{fig:probability_score_unseen10}, where we find that our approach well preserves the superior disentanglement ability in terms of in- and out-of-distribution samples across different open-set ratios. The performances for different random splits (R1 to R5) are shown in Fig.~\ref{fig:openset_score_dist_ablation_unseen}, where we observe that our approach keeps showing stable state-of-the-art performance over different open-set splits.

\section{Ablation for Noise Disturbance}
In this section, we introduce the OS-SAR performances with several most superior baselines and our approach under Gaussian noise disturbance on the NTU60~\cite{shahroudy2016ntu} dataset using CTRGCN~\cite{chen2021channel} backbone in Tab.~\ref{tab:noise}. Similar to the previous ablation in the supplementary, we keep using the four outstanding baselines, SoftMax, DEAR, ARPL, and RPL due to the limited time. A skeleton sequence can be denoted as $s\in \mathbb{R}^{3\times T \times N_j }$. Then we generate the Gaussian noise from a normal distribution, which could be denoted by $n\in\mathbb{R}^{3\times T \times N_j }$. Then the skeleton sequence with noise can be denoted as $s_{n} = s + \gamma * n$, where $\gamma$ is chosen as $0.3$. The Gaussian noise is added to both the training set and the testing set.
After adding Gaussian noise into the skeleton dataset, a clear performance decay is delivered for the baselines and our CrossMax, illustrating the negative effect of the noise disturbance for OS-SAR. However, CrossMax shows overall less performance decay when dealing with noise disturbance. Our CrossMax approach keeps delivering state-of-the-art performance under noise disturbance. Next, we showcase the predicted open-set probabilities for in- and out-of-distribution samples in Fig.~\ref{fig:probability_score_noise}, where the superior disentanglement ability of in- and out-of-distribution samples is delivered by CrossMax across different open-set ratios. The performances for different random splits (R1 to R5) are presented in Fig.~\ref{fig:openset_score_dist_ablation_noise}. Our approach delivers the most stable and state-of-the-art performances over different open-set splits for OS-SAR on NTU60~\cite{shahroudy2016ntu} cross-subject evaluation by using CTR-GCN~\cite{chen2021channel} as the backbone.

\section{Ablation under Occlusions}
In this section, we would like to introduce the experimental results of the OS-SAR with occlusion on NTU60~\cite{shahroudy2016ntu} cross-evaluation by using CTRGCN~\cite{chen2021channel} as the backbone in Tab.~\ref{tab:occ}.
To generate random occlusions, we randomly choose one occlusion ratio $\theta$ from $\{10\%, 20\%, 30\%\}$ to turn $\theta$ of the coordinates in a skeleton sequence as zeros, which can simulate the random occlusion. Occlusion is harmful to the OS-SAR since it may cause geometric discontinuity in the model, which makes the OS-SAR task quite challenging due to the sparsity of the skeleton data format. Compared with the same task conducted w/o random occlusion in Tab.~\ref{tab:occ}, all the baselines show obvious performance decays when dealing with occluded skeletons. When looking at the predicted open-set probabilities in terms of the in- and out-of-distribution samples in Fig.~\ref{fig:probability_score_occ}, we could find that more overlaps are presented, however, our CrossMax can preserve the great disentanglement ability for the in- and out-of-distribution samples. CrossMax can achieve $84.44\%$, $69.89\%$, and $88.69\%$ of O-AUROC, O-AUPR, and C-ACC for cross-subject evaluation and $86.30\%$, $74.07\%$, and $92.66\%$ of O-AUROC, O-AUPR, and C-ACC for cross-view evaluation respectively. We showcase the OS-SAR performances of the leveraged approaches under random occlusion for different splits in Fig.~\ref{fig:openset_score_dist_ablation_occ}, where we could observe that large performance discrepancies are showcased in all the leveraged OS-SAR baselines while our CrossMax shows the most stable performances.
\begin{table}[htb!]

\centering
\caption{Ablation study for OS-SAR under random occlusion disturbance using CTRGCN on NTU60~\cite{shahroudy2016ntu} dataset for cross-view and cross-subject evaluations, where results are averaged on five random splits.}
\label{tab:occ}
\scalebox{0.90}{\begin{tabular}{lllllll} 
\toprule\midrule
\multirow{2}{*}{\textbf{Method}} & \multicolumn{2}{l}{\textbf{O-AUROC}} & \multicolumn{2}{l}{\textbf{O\_AUPR}} & \multicolumn{2}{l}{\textbf{C-ACC}} \\ 
\cline{2-7}
 & \textbf{CS} & \textbf{CV} & \textbf{CS} & \textbf{CV} & \textbf{CS} & \textbf{CV} \\ 
 \midrule
 \multicolumn{5}{l}{\textbf{\circled{1} Without Occlusion.}} \\
\midrule
SoftMax & 83.68 & 87.77 & 67.37 & 76.38 & 90.56 & 93.83 \\
RPL& 84.02 & 88.06 & 67.86 & 76.75 & 90.82 & 95.38 \\
ARPL  & 84.13 & 88.37 & 68.24 & 76.58 & 91.00 & 95.45 \\
DEAR & 83.11 & 87.54 & 63.07 & 75.52 & 84.14 & 95.41 \\ 
\midrule
{\cellcolor{LightCyan}}  Ours &{\cellcolor{LightCyan}}\textbf{90.62} &{\cellcolor{LightCyan}}\textbf{94.14} &{\cellcolor{LightCyan}}\textbf{80.32} &{\cellcolor{LightCyan}}\textbf{88.07} & {\cellcolor{LightCyan}}\textbf{93.68} & {\cellcolor{LightCyan}}\textbf{97.51}  \\
\midrule
\multicolumn{5}{l}{\textbf{\circled{2} With Random Occlusions.}} \\
\midrule
SoftMax & 77.34 & 80.09 & 58.88 & 63.67 & 67.32 & 71.24 \\
RPL & 76.72 & 78.96 & 57.71 & 61.63 & 74.39 & 78.97 \\
ARPL & 79.92 & 79.55 & 61.55 & 63.35 & 87.18 & 88.01 \\
DEAR & 79.79 & 81.45 & 60.45 & 65.23& 87.44 & 90.75 \\ 
\midrule
{\cellcolor{LightCyan}}Ours & {\cellcolor{LightCyan}}\textbf{84.44}& {\cellcolor{LightCyan}}\textbf{86.30}& {\cellcolor{LightCyan}}\textbf{69.89}& {\cellcolor{LightCyan}}\textbf{74.07}& {\cellcolor{LightCyan}}\textbf{88.69}& {\cellcolor{LightCyan}}\textbf{92.66}  \\
\midrule\bottomrule
\end{tabular}}
\end{table}
\begin{table}[htb!]
\centering
\caption{Comparison with our implemented MM-ARPL on NTU60~\cite{shahroudy2016ntu} cross-subject evaluation on CTRGCN backbone, where the results are averaged among five random splits.}
\vskip-1ex
\label{tab:comparison_with_arpl_mm}
\resizebox{\linewidth}{!}{\begin{tabular}{l|lll} 
\toprule
\toprule
\textbf{Method} & \textbf{O-AUROC} & \textbf{O-AUPR} & \textbf{C-ACC} \\ 
\midrule
SoftMax~\cite{hendrycks2016baseline} & 83.68 & 67.37 & 90.56 \\
ARPL~\cite{chen2021adversarial} & 84.13 & 73.27 & 91.00 \\
Ensemble (MM-SoftMax) & 86.23 & 71.35 & 93.31 \\
MM-ARPL & 87.60 & 73.27 & 93.67 \\
\midrule
CrossMMD (Ours) & 88.31 & 74.80 & 93.68 \\
CrossMax (Ours) & \textbf{90.62} & \textbf{80.32} & \textbf{93.68}  \\
\midrule
\bottomrule
\end{tabular}}
\vskip-3ex
\end{table}
\begin{table*}[htb!]
\centering
\caption{Unseen classes for five random splits on NTU60~\cite{shahroudy2016ntu} dataset.}
\label{tab:app_ntu60}
\begin{tabular}{ll} 
\toprule\midrule
NTU60 & Unseen classes \\ 
\midrule\midrule
Run1 & 50, 40, 30, 37, 12, 48, 45, 49, 8, 29, 58, 13, 1, 39, 27, 47, 14, 52, 3, 44 \\
\midrule
Run2 & 41, 21, 52, 6, 12, 36, 24, 56, 35, 57, 15, 26, 39, 53, 19, 4, 27, 25, 17, 47 \\
\midrule
Run3 & 46, 10, 47, 39, 55, 14, 58, 53, 13, 40, 24, 9, 45, 23, 27, 3, 7, 54, 33, 17 \\
\midrule
Run4 & 21, 55, 11, 43, 41, 3, 52, 39, 46, 59, 47, 15, 17, 54, 40, 33, 9, 38, 31, 57 \\
\midrule
Run5 & 56, 14, 17, 7, 40, 52, 37, 50, 36, 6, 44, 11, 41, 9, 47, 24, 53, 2, 10, 58 \\
\midrule\bottomrule
\end{tabular}
\end{table*}
\begin{table*}[htb!]
\caption{Seen classes for five random splits on NTU120~\cite{liu2020ntu} dataset.}
\label{tab:app_ntu120}
\centering
\resizebox{\linewidth}{!}{
\begin{tabular}{l|l} 
\toprule\midrule
NTU120 & Seen classes \\ 
\midrule\midrule
Run1 & 0,  37,  52,  70,  96,  92,  91,   4,  39,  12,  46,  81,  87,  31,  72,  48,  16,  62, 42, 102, 112,  68,  56,  49,  22,  11,  88, 107,  93, 43 \\
\midrule
Run2 & 17,  90,  47,  80,  79,  48,  27,  82,  61,  53,  96, 117,  62,  35,  23,  85,   8,  98, 104, 77, 51,  75,  56, 105,  54,  25,  18,  44,  40, 109 \\
\midrule
Run3 & 76,   9,  57,  59,   5,  51,  83, 104,  73,  27,  92,  72,  42, 111, 100,  67, 105,   4, 101,  12,  84, 119,  15,  33,  78,  62,  82,  24,  65, 108 \\
\midrule
Run4 & 48,  12,  26,  63,  20, 109,  80,  33,  79,  67, 100,   6,  24,  11,  76,  61,  10,  59, 0,  99,  19,   4,  90,  58,  28,  88,  44,  95,  72,  18 \\
\midrule
Run5 & 45,   0,  44,  13, 100,  14,  32,  72, 101,  17,  39,  63,  20,  56, 105,  71,  78,  73, 8,  99,  19, 115,  23,  54,  12, 109,  15,  37,  88,  18 \\
\midrule\bottomrule
\end{tabular}}
\end{table*}

\begin{table*}[htb!]
\caption{Unseen classes for five random splits on ToyotaSmartHome~\cite{dai2022toyota} dataset.}
\label{tab:app_tyt}
\centering
\resizebox{\linewidth}{!}{\begin{tabular}{l|l} 
\toprule\midrule
TYT & Unseen classes \\ 
\hline\midrule
\multirow{2}{*}{Run1} & 'Drink.Fromcup', 'Cook.Cleandishes', 'Laydown', 'Enter', 'Takepills', 'Walk', 'Usetablet', 'Cook.Usestove', 'Leave', 'Eat.Snack', \\
 & 'Maketea.Boilwater', 'Cook.Cut', 'Pour.Frombottle', 'Drink.Fromglass', 'Uselaptop', 'WatchTV', 'Pour.Fromkettle', 'Usetelephone' \\
 \midrule
\multirow{2}{*}{Run2} & 'Leave', 'Usetelephone', 'Maketea.Boilwater', 'Cook.Usestove', 'Eat.Snack', 'Cook.Cleanup', 'Pour.Fromkettle', 'Cook.Stir', 'Walk', \\
 & ~'Usetablet',~'Pour.Frombottle',~'Drink.Fromglass',~'Getup',~'Makecoffee.Pourgrains',~'Drink.Fromcup',~'Takepills',~'Makecoffee.Pourwater',~'Cutbread' \\
 \midrule
\multirow{2}{*}{Run3} & 'Usetelephone', 'Makecoffee.Pourwater', 'Cook.Usestove', 'Maketea.Insertteabag', 'Uselaptop', 'Enter', 'Maketea.Boilwater', 'Cutbread', 'Pour.Frombottle', \\

 & ~ ~'Drink.Fromcan',~'Cook.Stir',~'Laydown',~'Cook.Cleanup',~'Drink.Fromcup',~'Readbook',~'Drink.Frombottle',~'Leave',~'Pour.Fromcan' \\
 \midrule
\multirow{2}{*}{Run4} & 'Cutbread', 'Usetelephone', 'Drink.Frombottle', 'Walk', 'Usetablet', 'Cook.Cleanup', 'Drink.Fromcan', 'Drink.Fromglass', 'Drink.Fromcup', 'Pour.Fromcan',~ \\

 & 'Makecoffee.Pourgrains',~'Maketea.Boilwater',~'Leave',~'Cook.Stir',~'Makecoffee.Pourwater',~'WatchTV',~'Laydown',~'Eat.Attable' \\
 \midrule
\multirow{2}{*}{Run5} & 'Enter', 'Eat.Attable', 'Pour.Frombottle', 'Eat.Snack', 'Cook.Cleanup', 'Takepills', 'Pour.Fromkettle', 'Sitdown', 'Makecoffee.Pourgrains', 'WatchTV',~ \\

 & 'Uselaptop',~'Drink.Frombottle',~'Drink.Fromcan',~'Cook.Cut',~'Readbook',~'Cutbread',~'Maketea.Boilwater',~'Maketea.Insertteabag' \\
\midrule\bottomrule
\end{tabular}}
\end{table*}

\section{Comparison with ARPL under Three Modalities}
Since our methods rely on three modalities derived from skeleton data, there would be a question regarding the comparison between our CrossMax with the best baseline implemented into the multi-modality setting, we showcase these experiments in Tab.~\ref{tab:comparison_with_arpl_mm}, where experiments are conducted on the NTU60~\cite{shahroudy2016ntu} cross-subject evaluation by using the CTRGCN~\cite{chen2021channel} as the backbone. ARPL~\cite{chen2021adversarial} is chosen due to its superior performance on the NTU60 dataset by using CTRGCN as the backbone. We train ARPL separately on three modalities, \eg, joints, bones, and velocities while averaging the outputs before the calculation of the final open-set probability, which is named MM-ARPL in Tab.~\ref{tab:comparison_with_arpl_mm}.  Apart from the MM-ARPL, we also report the performances of multi-modal SoftMax, noted as Ensemble, the vanilla SoftMax~\cite{hendrycks2016baseline}, and the vanilla ARPL~\cite{chen2020learning} in Tab.~\ref{tab:comparison_with_arpl_mm}. Compared with the vanilla ARPL, MM-ARPL achieves performance improvements by $3.47\%$, $5.03\%$, and $2.67\%$ in terms of O-AUROC, O-AUPR, and C-ACC, illustrating the importance of using multiple modalities for OS-SAR. Our CrossMax keeps surpassing MM-ARPL by $3.02\%$ and $7.05\%$ in terms of O-AUROC and O-AUPR, while the close-set performance also harvests benefits, showing the superior design of our approach when dealing with open-set challenges. Note that, O-AUROC and O-AUPR are the most important metrics to measure the open-set performance of the model, while the C-ACC is the minor metric in our benchmark to measure the close-set classification performance.

\section{Stability for Different Splits across Backbones}
We show the experimental results for OS-SAR by using different splits on NTU60 cross-subject evaluation across backbones in Fig.~\ref{fig:backbone}, where O-AUROC and O-AUPR evaluation results are presented. Compared with other approaches, our CrossMax achieves state-of-the-art performances overall while maintaining relatively regular shapes across different splits, demonstrating the superior generalizability of our approach.

\section{Evaluation Protocols}
More introductions regarding our evaluation protocols are shown next. We observe that by using different splits, the OS-SAR difficulties are not unified.
To ensure a fair comparison, we first randomly generate five splits for each dataset while fixing the classes for seen and unseen sets when a comparison is conducted, which guarantees fairness for future works. Considering the NTU60 dataset, we report the leveraged unseen classes in Tab.~\ref{tab:app_ntu60}, where the class index follows the category index setting from NTU60~\cite{shahroudy2016ntu}. The rest, not reported classes, serve as seen classes.
Considering the NTU120 dataset, we report the leveraged seen classes in Tab.~\ref{tab:app_ntu120}, where the class index follows the category index setting from NTU120~\cite{liu2020ntu}. The rest, not reported classes, serve as unseen classes.
Considering the ToyotaSmartHome dataset, we report the leveraged unseen classes in Tab.~\ref{tab:app_tyt}. The rest, not reported classes, serve as seen classes.
We use the training set samples from the seen classes during training while using the test set samples from both seen and unseen classes for testing. For the leveraged O-AUPR and O-AUROC metrics, we evaluate the approaches on the concatenation of the unseen test set and the seen test set. For the C-ACC metric, we evaluate the leveraged approaches in the OS-SAR benchmark only on the seen test set.

We use \underline{open-set probability} to refer to the probability associated with recognizing or categorizing data that belong to known classes but might also contain examples that do not belong to any of the known classes. The open-set probability is expected to be $1.0$ when the sample comes from the seen categories while it is expected to be $0.0$ when the sample comes from the unseen categories. In other words, it deals with situations where a classification model encounters data that is outside its trained classes.
Then we will give more introduction regarding the O-AUROC and O-AUPR metrics.

\noindent\textbf{O-AUROC:} The Area Under the Receiver Operating Characteristic (AUROC) curve, commonly serves as a metric in binary classification, which can perfectly measure the quality of the open-set probability following PMAL~\cite{lu2022pmal}, since the estimation of the open-set probability should be $1.0$ when the model is very certain about that the sample is from seen classes while the estimation of the open-set probability should be $0.0$ when the model is very certain about that the sample is from unseen classes. 
Thereby, the annotations for the unseen test set are all set to $0$ while the annotations for the seen test set are set to $1$ for the open-set evaluation. O-AUROC is leveraged to measure the performance of the model according to the plot of the true positive rate against the false positive rate by using various thresholds and calculating the area under this curve.
We first need to calculate the True Positive Rate (TPR) via Eq.~\ref{eq:tpr}, which can be regarded as sensitivity or recall.
\begin{equation}
\label{eq:tpr}
    TPR = \frac{TP}{TP+FN},
\end{equation}
where TP indicates the True Positives and FN indicates the False Negatives. 
Then we need to calculate the False Positive Ratio (FPR) for each threshold setting to measure the correctness of the probability. FPR can be calculated by Eq.~\ref{eq:fpr}, 
\begin{equation}
\label{eq:fpr}
    TPR = \frac{FP}{FP+TN},
\end{equation}
where FP indicates the False Positives and TN indicates the True Negatives. After that, we only need to plot the ROC curve using the TPR against the FPR for different thresholds. Then we can calculate the area under the ROC curve as the O-AUROC result.

\noindent\textbf{O-AUPR:} The Area Under the Precision-Recall (AUPR) curve is another well-established metric for evaluating the performance of binary classification models, especially considering class imbalances. The trade-off between precision and recall across different classification thresholds is measured. To obtain AUPR, first, we need to calculate the Precision and Recall as follows,

\begin{equation}
    Precision = \frac{TP}{TP + FP}
\end{equation}

\begin{equation}
    Recall = \frac{TP}{TP + FN}
\end{equation}
The predicted probabilities should be sorted in descending order. Then we can calculate the precision-recall curve by using various thresholds. Finally, we calculate the Area Under the Precision-Recall (AUPR) curve. 
This can also be done using numerical integration methods.
AUPR is particularly useful when handling imbalanced datasets, where negative instances are more than positive instances. It provides insight into the capability of the model to achieve the positive class correctly, especially in cases where the class of interest is rare and requires to be recognized with high precision. We name it O-AUPR where O indicates open-set.

\end{document}